\pdfoutput=1

\documentclass[11pt]{article}

\usepackage{EMNLP2022}

\usepackage{times}
\usepackage{latexsym}

\usepackage[T1]{fontenc}

\usepackage[utf8]{inputenc}

\usepackage{microtype}

\usepackage{inconsolata}
\usepackage{multirow}
\usepackage{booktabs}
\usepackage{graphicx}
\usepackage{tikz}
\usetikzlibrary{shapes, arrows,positioning,fit,backgrounds}
\tikzstyle{connector}=[draw,line width=1mm,-latex']
\tikzstyle{a}=[rectangle,thick,text centered,draw,align=center]
\newcommand{\tabitem}{~~\llap{\textbullet}~~}

\title{LawngNLI: A Long-Premise Benchmark for In-Domain Generalization from Short to Long Contexts and for Implication-Based Retrieval}

\author{William Bruno \and Dan Roth \\
	University of Pennsylvania \\
	\texttt{\{wwbruno, danroth\}@seas.upenn.edu} \\}

\begin{document}
\maketitle
\begin{abstract}
	
Natural language inference has trended toward studying contexts beyond the sentence level. An important application area is law: past cases often do not foretell how they apply to new situations and implications must be inferred. This paper introduces LawngNLI, constructed from U.S. legal opinions with automatic labels with high human-validated accuracy. Premises are long and multigranular. Experiments show two use cases. First, LawngNLI can benchmark for in-domain generalization from short to long contexts. It has remained unclear if large-scale long-premise NLI datasets actually need to be constructed: near-top performance on long premises could be achievable by fine-tuning using short premises. Without multigranularity, benchmarks cannot distinguish lack of fine-tuning on long premises versus domain shift between short and long datasets. In contrast, our long and short premises share the same examples and domain. Models fine-tuned using several past NLI datasets and/or our short premises fall short of top performance on our long premises. So for at least certain domains (such as ours), large-scale long-premise datasets are needed. Second, LawngNLI can benchmark for implication-based retrieval. Queries are entailed or contradicted by target documents, allowing users to move between arguments and evidence. Leading retrieval models perform reasonably zero shot on a LawngNLI-derived retrieval task. We compare different systems for re-ranking, including lexical overlap and cross-encoders fine-tuned using a modified LawngNLI or past NLI datasets. LawngNLI can train and test systems for implication-based case retrieval and argumentation.

\end{abstract}

\section{Introduction}
\label{sec:introduction}

This work proposes a new natural language inference (NLI) benchmark LawngNLI constructed from U.S. legal opinions via the Caselaw Access Project (\citealp{PresidentUn18}) that have been largely cleaned of in-line citations in order to read more naturally.\footnote{Code for obtaining LawngNLI and unfiltered-LawngNLI2 to be released at \url{http://cogcomp.org/page/publication_view/990}. LawngNLI contains about 140 thousand twinned examples. Our script and coded ID files should reconstruct LawngNLI using 1 CPU without GPUs in under 1 day (Intel Xeon Gold 6130 CPU @ 2.10GHz). A dataset from an earlier stage in construction process, unfiltered-LawngNLI2 (not studied in the paper), contains about 10.3 million untwinned candidate examples. Another script should reconstruct unfiltered-LawngNLI2 without GPUs by multiprocessing (about 3 days or fewer with 35 CPUs).} It follows the general three-label NLI formulation: given a pair of texts (premise and hypothesis), the goal is to predict the label of whether the premise entails, is neutral toward, or contradicts the hypothesis. LawngNLI's premises are especially long and are multigranular. Its automatic labels derive from the dataset construction using (negation-based) contradiction and (similarity-based) neutralization algorithms. These labels exhibit an accuracy of 88.8\% (94.7\% for high-confidence human labels) on a subset with human-validated gold labels. Examples are derived from actual inferences from past cases that judges wrote to apply usefully to new situations, relying on their expertise while avoiding crowdsourced labeling.

We conduct two sets of experiments on LawngNLI. First, regarding in-domain generalization to long contexts, we compare top long-sequence and short-sequence NLI models on LawngNLI. Models are transferred with and without intermediate fine-tuning on existing NLI benchmarks. We find that absent fine-tuning directly, models fall substantially short of top performance on our long premises. This continues to hold true when we leverage LawngNLI's multigranularity to control for domain shift, by further fine-tuning models on our own short premises. Top performance on our long premises is by short-sequence models prepended with a standard retrieval method (BM25 (\citealp{RobertsonZa09})) filtering across each premise. Absent their fine-tuning that uses long premises as inputs, however, these models would underperform on long premises at inference time. We provide evidence that this gap is likely robust to any deviations between the automatic and gold labels. Thus a large-scale long-premise dataset like LawngNLI is needed.

Second, regarding legal retrieval systems, we conduct a comparison of leading models by Recall@k for selecting target cases via implication/NLI-based retrieval (when a user provides arguments entailed or contradicted by target cases as queries), in our domain. Implication-based retrieval is an underexplored subtask (see, e.g., \citealp{SCBFM22}). Our zero-shot panel comprises Sentence-Transformers (\citealp{ReimersGu21}) lightweight bi-encoders pretrained using short-sequence NLI, user web queries, and general semantic relatedness. While baseline models transfer reasonably to our domain zero shot, we further compare re-ranking with models fine-tuned using several previous NLI datasets along with an adjusted retrieval version of LawngNLI. Future improvements can put more evidence within the range of human users' cognitive reach beyond the top result, including for law which could help make legal work more affordable and equitably accessible (see Section~\ref{sec:ethics}).

Overall, our main contributions are: \newline (1) A new NLI benchmark with multigranular premises much longer than in most existing NLI benchmarks across percentiles (see Table~\ref{tab:table1}). \newline (2) A benchmarking of models' ability to generalize from short context to long context on the same domain and examples. It shows that (in our domain at minimum) short-premise NLI models fall substantially short of top performance on long-premise NLI at inference time, unless a large-scale long-premise dataset is created to fine-tune on. \newline (3) A benchmarking of leading retrieval models on case retrieval with entailed or contradicted arguments as queries, comparing lexical overlap and fine-tuning using a modified LawngNLI or previous NLI datasets. LawngNLI provides a benchmark for future implication-based case retrieval systems.

\section{Significance for NLI and for Law}
\label{sec:significance}

\begin{table*}
	\centering
	{\scriptsize
		\begin{tabular}{p{2cm}|p{13cm}}\toprule
			\multicolumn{2}{c}{\textbf{Sample twin Entail/Contradict examples with same premise from LawngNLI}}  \\
			\midrule
			\multirow{2}{1.9cm}{Twin hypotheses with same premise, from ``analysis'' subset} & \tabitem \textit{Contradict:} city acted affirmatively to create or increase risk of harm on city street by ignoring residents' requests to reduce speed limit or by taking down residents' signs indicating drivers should adhere to a lower speed limit \\
			& \tabitem \textit{Entail:} city did not act affirmatively to create or increase risk of harm on city street by ignoring residents' requests to reduce speed limit or by taking down residents' signs indicating drivers should adhere to a lower speed limit \\
			\midrule
			\multirow{2}{1.9cm}{Additional hypotheses with same premise} & \tabitem \textit{Entail:} failing to enforce or lower the speed limit on a residential street ``did not create a `special danger' to a discrete class of individuals..[ed.: excerpted]..as opposed to a general traffic risk to pedestrians and other automobiles'' \\
			& \tabitem \textit{Contradict:} traffic laws and enforcement practices did not pose ``a general traffic risk to pedestrians and other automobiles'' \\
			\midrule
			Relevant excerpts of shared premise & \tabitem [ed.: Plaintiffs] ...submit that the City of Fort Thomas..violated their son's substantive due process rights by failing to act upon their request (and the requests of others) to lower the speed limit on the street..The police also removed signs posted by residents indicating that drivers should adhere to a 15 mile-per-hour speed limit..\\
			& \tabitem [ed.: Plaintiffs] ...alleged that the City's failure to maintain safe conditions on Garrison Avenue violated their son's substantive due process rights..established a ``state-created danger'' under DeShaney..\\
			& \tabitem ...DeShaney's holding..precludes [ed.: Plaintiffs'] argument that the Due Process Clause constitutionalizes a locality's choices about what speed limit to adopt for a given street or how to enforce that speed limit..\\
			& \tabitem There are two exceptions to the DeSha-ney rule..Under the second exception..a plaintiff may bring a substantive due process claim by establishing (1) an affirmative act by the State that either created or increased the risk that the plaintiff would be exposed to private acts of violence..\\
			& \tabitem [ed.: Plaintiffs] fail to satisfy any of the three requirements for establishing our circuit's ``state-created danger'' exception to DeShaney. First, the creation of a street and the management of traffic conditions on that street are too attenuated and indirect to count as an ``affirmative act''..\\
			\midrule
			Distractor excerpts of same premise & \tabitem ...After all, the City was told about the risks of not lowering the speed limit to 15 miles per hour (more accidents); it intentionally chose not to heed this warning (taking on the risk of more accidents); and the alleged risk came to pass when..was killed (an accident)..\\
			& \tabitem ...For in one sense, it could be said that all governing bodies act with deliberate indifference when they consider and reject a traffic-safety proposal of this sort that comes with known risks..\\
			\bottomrule
		\end{tabular}
	}
	\caption{\label{tab:sample1}
		Sample twin Entail/Contradict examples from LawngNLI. See Appendix Tables~\ref{tab:sample1a} and \ref{tab:sample2a} for detailed view.}
\end{table*}

Regarding in-domain generalization to long contexts, our work stands within a fast-growing research area on how models can learn to reason over long text. Benchmarks for NLI, or Recognizing Textual Entailment (RTE), stretch back to \citet{DaganGlMa05}. Recently, various ``efficient'' Transformer architectures have been proposed to address the obstacle of quadratic self-attention complexity in scaling to long sequences (\citealp{TDBM20}). LawngNLI's long premises frequently exceed the limits of such long-sequence models with efficient attention mechanisms, although on our long premises not exceeding those limits, the long-sequence models included here are outperformed by short-sequence models both with and without filtering of premise paragraphs by relevance to the hypotheses. Most existing NLI benchmarks, meanwhile, contain largely short premises. The recent ContractNLI (\citealp{KoreedaMa21}) is an exception, containing premises with lengths similar to ours though with a much smaller number of examples in a different domain (607 contracts as premises, paired with each in a common set of 17 shared hypotheses).\footnote{Two other outliers are two-label DocNLI (\citealp{YinRaXi21}) and three-label ConTRoL (\citealp{LCLZ21}). However, while their premises often exceed the usual 512 maximum sequence length, unlike LawngNLI they still largely are not near the typical maximum sequence lengths of key current long-sequence pretrained models (see Table~\ref{tab:table1}).}

Regarding legal retrieval systems, law is an important area where humans perform long-context NLI in practice. The core of legal advocacy is articulating what existing applicable law (cases, legislation, regulations, etc.) implies for a new situation. Practitioners must move between case text and the entailed and contradicted arguments that they aim to support or counter. Automatic systems are developed to help find relevant precedents, which requires filtering the millions of U.S. cases.

\section{LawngNLI Dataset}
\label{sec:dataset}

We construct LawngNLI beginning with all citations with parentheticals in U.S. state and federal case opinions, via the Caselaw Access Project (\citealp{PresidentUn18}). Its hypotheses derive from the actual inferences from past cases applied by judges to decide between competing parties' positions across a large cross-section of U.S. opinions. Thus we hope its examples capture the variety of reasoning underlying the useful arguments that legal practitioners aim to support or undermine during their daily work.

\subsection{Construction Steps}

Key steps are outlined below. They are presented in Figure~\ref{fig:construction} and further detailed in Appendix Section~\ref{sec:datasetconstruction}. 

When judges cite another case in an opinion, they may highlight content or takeaways from that case in a parenthetical which we use as an initial hypothesis.\footnote{These explanatory parentheticals are used by, for example, the legal research platform Casetext (\citealp{Arredondo17}).} 

\begin{enumerate}
\item We begin with Entail examples: long premises are the majority opinion cited alongside the parenthetical, and short premises are the cited opinion pages (extracted using Eyecite (\citealp{CushmanDaLi21})). Filters screen for hypotheses that are not flagged as conflicting with premises and are valid inputs for our later contradiction algorithm.
\item Stratified by the Cartesian product of geography and pivotal negation versus not, we apply a contradiction algorithm and a neutralization algorithm to convert a random 1/3 each of the remaining Entail examples into Contradict and Neutral examples, respectively. The contradiction algorithm removes or (if absent) adds pivotal negation in the hypotheses assigned Contradict, by modifying a method from \citet{BiluHeSl15}). All examples are filtered for NLI difficulty. Then the neutralization algorithm re-pairs the hypotheses assigned Neutral with alternative similar premises not adjacent in the citation network.
\item The dataset is rebalanced on labels and pivotal negation, citation spans are removed from the premises so that they read more naturally, and long premises are prepended with copied paragraphs from the end (the minimum number including at least 512 tokens) to limit models from relying on cues near the start of the underlying opinion.
\end{enumerate}

\begin{table*}
	\centering
	{\scriptsize
		\begin{tabular}{@{}c|c@{}c@{}c@{}c@{}}\toprule
			\multirow{2}{*}{\centering \textbf{LawngNLI}}&\multicolumn{2}{c}{\textbf{Long premises}}&\multicolumn{2}{c}{\textbf{Short premises}}\\
			\cmidrule{2-3} \cmidrule{4-5}
			&\textbf{``Analysis'' subset}&\textbf{Full}&\textbf{``Analysis'' subset}&\textbf{Full}\\
			\midrule
			\multirow{2}{3cm}{\centering Premise length}&\multirow{2}{3cm}{\centering [970, 1527, 2339, 3154, 3693]}&\multirow{2}{3cm}{\centering [1285, 2179, 3692, 6044, 9238]}&\multirow{2}{3cm}{\centering [301,  462,  711,  925, 1397]}&\multirow{2}{3cm}{\centering [331,  498,  746,  966, 1581]}\\
			&&&&\\
			Hypothesis length&22&21&22&21\\
			Hypothesis negation&[0.579, 0.583, 0.586]&[0.574, 0.578, 0.583]&[0.579, 0.583, 0.586]&[0.574, 0.578, 0.583]\\
			Unique training premises&27246&47803&29004&52643\\
			Unique training hypotheses&69545&124486&69545&124486\\
			Training examples&71442&128520&71442&128520\\
			&&&&\\
			Text sources&\multicolumn{4}{c}{Legal case opinions}\\
			\midrule
			\textbf{Existing datasets}&\textbf{ContractNLI}&\textbf{anli}&\textbf{DocNLI}&\textbf{ConTRoL-dataset}\\
			\midrule
			\multirow{2}{3cm}{\centering Premise length}&\multirow{2}{3cm}{\centering [871, 1281, 2019, 2741, 4134]}&\multirow{2}{3cm}{\centering [14, 28, 63, 80, 95]}&\multirow{2}{3cm}{\centering [57,   73,  115,  557, 1050]}&\multirow{2}{3cm}{\centering [55.6,  138,   333,   996,  1147]}\\
			&&&&\\
			Hypothesis length&19&14&57&16\\
			Hypothesis negation&[0.284, 0.391, 0.012]&[0.074, 0.069, 0.197]&[0.187, 0.202]&[0.094, 0.078, 0.107]\\
			Unique training premises&423&377440&488673&1530\\
			Unique training hypotheses&17&1168146&572368&6238\\
			Training examples&7191&3233665&942314&6719\\
			&&&&\\
			Text sources&Contracts&\multirow{3}{3cm}{\centering Wikipedia, news, etc, SNLI, MNLI, NLI version of FEVER}&\multirow{3}{3cm}{\centering ANLI, SQuAD, CNN/DailyMail, DUC (2001), Curation}&\multirow{3}{3cm}{\centering Various civil service exams}\\
			&&&&\\
			&&&&\\
			\bottomrule
		\end{tabular}
	}
	\caption{\label{tab:table1}
		Descriptive statistics of NLI datasets. Negation words [`no',`not',`never',`none',`nobody',`nothing', `neither',`nor',`cannot'] or contains ``n't''. Proportions are by label: Entail/Neutral/Contradict or Entail/Not entail. \textit{About 50\% of LawngNLI's hypotheses contain \textit{pivotal} negation}, even though over 50\% contain negation under the keyword definition (used here for comparability across datasets). See Appendix Section~\ref{sec:datasetconstruction} on dataset construction (the twinning step leads to the larger number of unique hypotheses than unique premises). Token lengths are [10, 25, 50, 75, 90] percentiles or an average via a RoBERTa (\citealp{LiuOtGoDuJoChLeLeZeSt19}) tokenizer.}
\end{table*}

Table~\ref{tab:sample1} shows sample examples from our dataset. Descriptive statistics (Table~\ref{tab:table1}) show that its long premises skew much longer than premises in other key existing NLI datasets.

\subsection{Automatic Labels and Human Assessment}
\label{sec:labels}

LawngNLI includes only automatic NLI labels. The Entail labels were effectively ``annotated'' by each judge authoring the (hypothesis) parenthetical citing another case's pages, but our algorithms constructing Neutral and Contradict examples could import some error rate. So with a balanced sample of long-premise and short-premise examples without prepending, these labels were assessed for accuracy. 

Screened Amazon Mechanical Turk workers provided 300 gold labels for LawngNLI examples. Because our long premises are lengthy, there is a particular risk of partly random guessing by workers. Even small frequencies can erroneously multiply the estimated error rates of our labels. As such, for each example, two workers independently chose a label and confidence level (``probably'' or ``definitely''). Gold labels were adopted on examples where both workers chose the same label (unanimity). Where the workers chose different labels, no label carries unanimous confidence and the example is not included. This process continued until 300 gold labels were obtained. Detailed steps are outlined in Appendix Section~\ref{sec:humanassessment}.

We find a 88.8\% human-validated accuracy (94.7\% for high-confidence labels, where both workers chose ``definitely''). Table~\ref{tab:humanassessment} shows human-assessed accuracies of our automatic labels.

To better understand where dataset errors arise, in Appendix Table~\ref{tab:label} we present several examples where automatic and gold labels differ. In the first, the citation parser linked the wrong citation to the parenthetical underlying the hypothesis. So the example was incorrectly automatically labeled as Entail, while workers correctly labeled the example as Neutral. In the second example, the hypothesis is the conjunction of two possible conditions for a non-medical source to be given the weight of an acceptable medical source under the applicable regulations. Yet along with the first condition, the premise describes the opposite of the first condition. Thus the automatic label Contradict is arguably correct, while the worker label Entail may have arisen if both workers saw that the second condition is correct while misassessing the first condition. And in the third example, the premise and hypothesis were re-paired by our neutralization algorithm and so automatically labeled Neutral. However, despite safeguards such as excluding majority opinions from adjacent cases in the citation network as candidate premises, the workers recognized that the premise arguably contradicts the hypothesis despite deriving from a case not cited by the hypothesis's underlying parenthetical. Thus the automatic label is incorrect and the worker label is correct.

\begin{table}
	\centering
	{\footnotesize
		\scalebox{0.7}{
			\begin{tikzpicture}[node distance = 2.2cm,show background rectangle,inner frame sep=0mm]
				\node [a, text width=4cm] (case) {Premise, from cited case:\\ \vspace{\baselineskip} \textit{A v. B}\\ \textbf{Majority opinion (long version), containing cited pages (short version)}\\ \textit{Minority opinions}};
				\node [a, text width=4cm, right=of case] (parenthetical) {Hypothesis, from\\ citing parenthetical:\\ \vspace{\baselineskip} \textit{C contradicted by case}\\ \textit{A v. B at 2-3} (\textbf{C always false}).};
				\begin{scope}[on background layer]
					\node [a, minimum width=9cm, fit=(case) (parenthetical), inner xsep=1mm, inner ysep=4mm] (2) {};	
					\node[anchor=north] at (2.north) {\textbf{1. Extract Entail examples, filter for validity/invertibility.}};
					\node[a, minimum width=9cm, anchor=south] at (2.north) {\textbf{Source}: Millions of U.S. state and federal case opinions, via Caselaw Access Project};
				\end{scope}
				\node [a, text width=1.5cm, below=3.5cm of 2] (3b) {{\footnotesize\textbf{Entail}}};
				\node [a, text width=1.5cm, left=of 3b] (3a) {{\footnotesize\textbf{Neutral}}};
				\node [a, text width=1.5cm, right=of 3b] (3c) {{\footnotesize\textbf{Contradict}}};
				\path [connector] (parenthetical) -- (case);
				\path [connector] (2) -- (3a);
				\path [connector] (2) -- (3b);
				\path [connector] (2) -- (3c);
				\node [a, fill={rgb,1:red,1;green,1;blue,1}, text width=10.5cm, below=1cm of 2] (difficulty) {\textbf{Filter for difficulty}};
				\node[text height=0.8cm,text width=3cm, anchor=south] at (difficulty.north) (blank) {};
				\node[a, fill={rgb,1:red,1;green,1;blue,1}, text width=4cm, anchor=south east] at (difficulty.north east) {\textbf{Invert between pivotal negation/non-negation}};
				\node[a, fill={rgb,1:red,1;green,1;blue,1}, text width=4cm, anchor=north west] at (difficulty.south west) {\textbf{Re-pair hypothesis to maximum-similarity premise not adjacent in citation network}};
				\node [a, text width=9cm, below=0cm of 3b] (unfiltered) {\textbf{2. Neutralization and contradiction algorithms above,\\ to obtain unfiltered\_LawngNLI}=\\ 4.8 million untwinned candidate examples};
				\node [a, text width=9cm, below=0cm of unfiltered] (balancing) {\textbf{3. Balancing on NLI Label and Pivotal Negation}\\ \textbf{(Downsampling and Twinning)}};
				\node [a, text width=9cm, below=0cm of balancing] (citation) {\textbf{4. Citation Removal, Premise Prepending, and Dataset Split}};
				\node [a, text width=9cm, below=0cm of citation] {\textbf{LawngNLI}};
				\node[anchor=south west] at (blank.north west) {\textbf{1/3}};
				\node[anchor=south] at (blank.north) {\textbf{1/3}};
				\node[anchor=south east] at (blank.north east) {\textbf{1/3}};
			\end{tikzpicture}
		}
	}
	\caption{\label{fig:construction}
		Major steps in LawngNLI dataset construction process. Steps are detailed in Appendix Section~\ref{sec:datasetconstruction}.}
\end{table}

\section{Experimental Evaluation}
\label{sec:experimentalevaluation}

Our experiments demonstrate two applications of LawngNLI.\footnote{For all experiments, we use only LawngNLI's ``analysis'' subset: with long premises at most 4096 tokens, via a RoBERTa (\citealp{LiuOtGoDuJoChLeLeZeSt19}) tokenizer.} See implementation details in Appendix Section~\ref{sec:implementation}.

\noindent \textbf{1. Generalization} First, we evaluate whether models perform competitively on LawngNLI's long premises, before and after fine-tuning on existing NLI benchmarks and/or its short premises (with LawngNLI's multigranularity, we can evaluate on long premises within the same domain). 

\textbf{RQ1:} Can top NLI models approach top performance on LawngNLI with long premises, absent fine-tuning directly with our long premises?

\textbf{RQ1A:} Is our answer to RQ1 robust to any error rate in LawngNLI's automatic labels?

\noindent \textbf{2. Retrieval} Second, we evaluate retrieval and NLI models on a retrieve-and-re-rank approach to NLI-based (entailed or contradicted arguments as queries) case retrieval. 

\textbf{RQ2:} How do models compare on implication-based case retrieval?

\begin{table}
	\centering
	{\scriptsize
		\begin{tabular}{@{}p{2.2cm}@{}|cccc@{}}\toprule
			\multirow{2}{2.1cm}{\centering \textbf{LawngNLI automatic labels}}&\multicolumn{2}{c}{\textbf{Short premise}}&\multicolumn{2}{c}{\textbf{Long premise}}\\
			\cmidrule{2-3} \cmidrule{4-5}
			& \textbf{All} & \textbf{Negation} & \textbf{All} & \textbf{Negation}\\
			\hline
			\multicolumn{5}{c}{Agreed-upon (gold) labels}\\
			\hline
			\centering Accuracy  &0.92&0.901&0.888&0.87\\
			\centering N &160&76&140&66\\
			\hline
			\multicolumn{5}{c}{\textit{High-confidence} agreed-upon (gold) labels}\\
			\hline
			\centering Accuracy &0.972&0.976&0.947&0.905\\
			\centering N &81&39&68&31\\
			\hline
			\multicolumn{5}{c}{Full assessment set}\\
			\hline
			\centering Worker agreement &0.758&0.71&0.761&0.75\\
			\multirow{2}{2.1cm}{\centering High confidence, if agreement}&0.506&0.513&0.486&0.47\\
			&  &  &  &  \\
			\centering N &211&107&184&88\\
			\bottomrule
		\end{tabular}
	}
	\caption{\label{tab:humanassessment}
		Human assessment of a stratified random sample of LawngNLI's \textit{``analysis'' subset} (sequence length of long premise at most 4096).  The split refers to \textit{pivotal} negation. Provided accuracies are balanced (macro-averaged recall). High-confidence labels are when the two workers who labeled a given example both clicked ``definitely'' (versus ``probably'') the label.}
\end{table}

\subsection{In-Domain Generalization to Long Contexts}
\label{sec:generalization}

To compare generalization from short to long contexts, we build a panel of 28 NLI models as follows:

\begin{enumerate}
\item We begin with 7 pretrained models that are top performing on key existing NLI benchmarks. HuggingFace (\citealp{WCDrSDMCFDS20}) model names with leaderboard positions on existing NLI benchmarks are listed in Appendix Section~\ref{sec:pretrained}.
\item To improve ability on general NLI, we add model versions with intermediate fine-tuning on other NLI benchmarks: three-label ANLI  (\citealp{NWDBWK20}) which contains MNLI (\citealp{WilliamsNaBo18}), three-label ConTRoL (\citealp{LCLZ21}),\footnote{Following this paper, we fine-tune on ANLI and then ConTRoL.} and two-label (Entail, Not Entail) DocNLI (\citealp{YinRaXi21}).\footnote{See Appendix Section~\ref{sec:implementation} about converting LawngNLI to two labels.}
\item We then fine-tune on LawngNLI by adapting the code from \citet{XZCTFLS21}.
\end{enumerate}

Past NLI benchmarks have included artifacts spuriously correlated with their labels. To check for LawngNLI, we fine-tune and evaluate our models on hypotheses and premises only in Appendix Table~\ref{tab:table11} (\citealp{GururanganSwLeScBoSm18}; \citealp{PNHRD18}; \citealp{Tsuchiya18}; \citealp{YinRaXi21}). Our labels show some modest predictability above random from our hypotheses at around 0.55 at the highest (in line with other NLI datasets\footnote{For hypotheses, similar to ANLI (\citealp{NWDBWK20}) A1 at 0.497 and MNLI at 0.55 (\citealp{WilliamsNaBo18}; \citealp{PNHRD18}) and slightly above ANLI later rounds and ConTRoL (\citealp{LCLZ21}) in the 0.40s.}) and from our long premises at 0.45 at the highest (slightly higher than another imperfectly symmetric benchmark (ConTRoL (\citealp{LCLZ21})) at 38.56\%) and short premises at around 0.52 at the highest.

Especially for short-sequence models, some difficulty on long premises may be due to distraction by less relevant text. We address this by adding a version of each model with a filtering step prepended, selecting the 5 paragraphs with highest BM25 (\citealp{RobertsonZa09}) score across the premise when querying the hypothesis. This filtering step allows our models to rival their performance on LawngNLI's short premises, when inputting only its unexcerpted long premises.

\subsubsection{RQ1: Are NLI Models  Competitive on LawngNLI's Long Premises, Absent Fine-Tuning with Our Long Premises?}

Among our NLI models fine-tuned with existing benchmarks only and/or LawngNLI with short premises, the top model trails the top model fine-tuned with our long premises in accuracy by 5 percentage points (bolded in columns (4) and (6) among all 21 three-label models in Appendix Table~\ref{tab:table32a}; the top 6 models among these on both our short and long premises plus all long-sequence models are shortlisted in Table~\ref{tab:table32}).\footnote{Appendix Tables \ref{tab:table2a} and \ref{tab:table2b} show minimal differences on examples with hypotheses with versus without pivotal negation and slightly lower performance on examples with above-median length premises.}

Thus these models fall substantially short of top performance on long-premise NLI, unless a large-scale long-premise NLI dataset is constructed and used for fine-tuning. Models have more to learn here from long contexts than short contexts alone teach.

Generalizing from short to long premises, we control for much example structure by keeping the same domain and examples.
However, judge excerpting could create some long versus short premise differences besides length, and it is possible that these differences are impacting our results. Such differences may be somewhat unavoidable when excerpting NLI premises that can cover many arguments. Relevant passages depend on the given hypothesis and so may be variably imbalanced relative to the long premise. In contrast to multilevel summaries, our motivation is comparing nested premises that read like premises naturally occurring in the domain at their respective lengths.

We would thus argue that differences arising from human excerpting need not confound drawing wider parallels to our results. Counterpart (if differently structured) differences between naturally occurring short contexts versus long contexts likely exist in various domains. These differences are arguably properly includable in our experiments since they impact model generalization from short to long contexts in such other domains as well. Still it remains to be studied if large-scale long-context datasets are needed to perform competitively on other long-context NLP tasks and domains.

\subsubsection{RQ1A: Is RQ1 Robust to Any Error Rate in LawngNLI's Automatic Labels?}
\label{sec:goldlabels}

On our human-validated subset, we check how the gap in balanced accuracies from RQ1 (i.e., between our top models fine-tuned using long premises versus those fine-tuned using short premises) is impacted by evaluating against our automatic labels, as compared to against our gold labels. While sampling variation leads to some divergence of this human-validated subset from the test set, this bias in balanced accuracies due to automatic labels (on our human-validated subset) provides a noisy estimate of the same bias for the test set.\footnote{The modest gold subset size provides low power for statistical testing, but we can still calculate with the point values.}

In Appendix Table~\ref{tab:table4}, we can check this impact of the automatic labels on the performance gap between the top models fine-tuned using long premises (column 6) versus top models fine-tuned using short premises (column 4) from Table~\ref{tab:table32}. For the single model (among these top 6) with the largest performance gap from Table~\ref{tab:table32}, this gap is biased on the human-validated subset by the automatic labels by -0.066. Averaging across all top 6 models, this bias is -0.016. These non-positive biases provide evidence that the performance gap from RQ1 likewise likely does not arise from bias due to the automatic labels.

\subsection{Implication-Based Legal Retrieval}
\label{sec:evaluation2}

Major legal retrieval systems (including leading commercial systems, according to publicly available information) rely on signals from lexical keywords, semantic similarity, and question answering (Section~\ref{sec:relatedwork}). But to our knowledge, such systems do not utilize signals from legal NLI. In contrast, LawngNLI's hypotheses are derived from the actual implications from the premise case that judges write within new cases when applying the premise case as precedent.

Suppose a user wished to retrieve documents that support or refute an argument: Can leading retrieval and NLI models perform well (zero shot or with fine-tuning) on retrieving the target document when given that argument as the query? We test this question by comparing models on implication-based retrieval in our domain. 

\subsubsection{RQ2: How Do Models Compare on Implication-Based Case Retrieval?}

\begin{table*}
	\centering
	{\scriptsize
		\begin{tabular}{@{}p{4.5cm}@{}|@{}p{0.1cm}@{}ccc@{}c@{}c@{}c@{}c@{}c@{}c@{}}\toprule
			\multirow{3}{4.5cm}{\centering \textbf{Needs long premises for fine-tuning}}&  &  &  &  &  &  &  &  &  &\\
			&  &  &  &  &  &  &  &  &  &\\
			&&\multicolumn{5}{c}{\textbf{No}}&&\multicolumn{3}{c}{\textbf{Yes}}\\
			\cmidrule{1-1} \cmidrule{3-7} \cmidrule{9-11}
			\multirow{3}{4.5cm}{\centering \textbf{Fine-tuning}}	 &
			\phantom{abc}&\multicolumn{3}{@{}c}{\multirow{3}{2.2cm}{\centering \textbf{Short premise}}}&
			\phantom{abc}&\multirow{3}{1.8cm}{\centering \textbf{BM25 retrieval on short premise}}&
			\phantom{abc}	 &\multirow{3}{1.8cm}{\centering \textbf{Long premise}}&
			\phantom{abc}&\multirow{3}{1.8cm}{\centering \textbf{Long premise filtered by BM25 retrieval}}\\
			&  &  &  &  &  &  &  &  &  &\\
			&  &  &  &  &  &  &  &  &  &\\
			\cmidrule{1-1} \cmidrule{3-5} \cmidrule{7-7} \cmidrule{9-9} \cmidrule{11-11}
			\multirow{4}{4.5cm}{\centering \textbf{Long premise filtered by BM25 retrieval at evaluation}} && \textbf{No} & \textbf{No} & \textbf{Yes} && \textbf{Yes} && \textbf{No} && \textbf{Yes}\\
			&  &  & \textbf{[512} & &  &  &  &  &  &\\
			&  &  & \textbf{[tokens]} & &  &  &  &  &  &\\
			&  & \textbf{(1)} & \textbf{(2)} & \textbf{(3)} &  & \textbf{(4)} &  & \textbf{(5)} &  & \textbf{(6)}\\
			\midrule
			\multicolumn{11}{c}{[Entail/Neutral/Contradict. Chance=1/3]}\\
			\midrule
			\centering albert-xxlarge-v2\_anli&&0.742+/-0.014&0.742+/-0.014&0.81+/-0.013&&0.817+/-0.012&&0.789+/-0.013&&0.853+/-0.011\\
			\centering albert-xxlarge-v2\_ConTRoL-dataset&&0.725+/-0.014&0.725+/-0.014&0.804+/-0.013&&0.806+/-0.013&&0.781+/-0.013&&0.851+/-0.011\\
			\centering albert-xxlarge-v2\_vanilla&&0.743+/-0.014&0.743+/-0.014&0.816+/-0.012&&\textbf{0.825+/-0.012}&&0.781+/-0.013&&0.852+/-0.011\\
			&&&&&&&&&&\\
			\centering allenai\_longformer-base-4096\_anli&&0.562+/-0.016&0.652+/-0.015&0.706+/-0.014&&0.707+/-0.014&&0.691+/-0.015&&0.784+/-0.013\\
			\centering allenai\_longformer-base-4096\_ConTRoL-dataset&&0.56+/-0.016&0.641+/-0.015&0.698+/-0.015&&0.691+/-0.015&&0.693+/-0.015&&0.781+/-0.013\\
			\centering allenai\_longformer-base-4096\_vanilla&&0.541+/-0.016&0.604+/-0.015&0.691+/-0.015&&0.701+/-0.015&&0.51+/-0.016&&0.791+/-0.013\\
			&&&&&&&&&&\\
			\centering facebook\_bart-large\_anli&&0.664+/-0.015&0.675+/-0.015&0.808+/-0.013&&0.809+/-0.013&&0.76+/-0.014&&0.868+/-0.011\\
			\centering facebook\_bart-large\_ConTRoL-dataset&&0.665+/-0.015&0.666+/-0.015&0.809+/-0.013&&0.814+/-0.012&&0.758+/-0.014&&0.868+/-0.011\\
			\centering facebook\_bart-large\_vanilla&&0.667+/-0.015&0.699+/-0.015&0.813+/-0.012&&0.807+/-0.013&&0.782+/-0.013&&\textbf{0.875+/-0.011}\\
			&&&&&&&&&&\\
			\centering google\_bigbird-roberta-base\_anli&&0.613+/-0.015&0.666+/-0.015&0.764+/-0.014&&0.774+/-0.013&&0.77+/-0.013&&0.829+/-0.012\\
			\centering google\_bigbird-roberta-base\_ConTRoL-dataset&&0.601+/-0.015&0.656+/-0.015&0.761+/-0.014&&0.769+/-0.013&&0.757+/-0.014&&0.819+/-0.012\\
			\centering google\_bigbird-roberta-base\_vanilla&&0.648+/-0.015&0.657+/-0.015&0.767+/-0.013&&0.777+/-0.013&&0.763+/-0.014&&0.827+/-0.012\\
			\midrule
			\centering Maximum of p-values versus (6) &&0&0&0&&0&&&&\\
			\midrule
			\centering N &  \multicolumn{10}{c}{3966} \\
			\bottomrule
		\end{tabular}
	}
	\caption{\label{tab:table32}
		Accuracy on LawngNLI's ``analysis'' subset \textit{with long premises}: Top 6 three-label models on both short and long premises plus long-sequence models (for all 28 models, see Appendix Table~\ref{tab:table32a}). The error provided is the larger deviation of the Clopper-Pearson (\citealp{ClopperPe34}) exact binomial 95\% confidence bounds. All p-values round to zero ($<$0.0005) from an exact binomial McNemar's (\citealp{McNemar47}) test for a statistically significant difference in accuracies between each version fine-tuning using short premises as inputs (1-4) and the best version fine-tuning using long premises as inputs (6). For (2), 512 tokens is the overall sequence limit.
	}
\end{table*}

\begin{table*}
	\centering
	{\scriptsize
		\begin{tabular}{@{}p{4.5cm}@{}|@{}p{0.1cm}ccccccc}\toprule
			\multirow{3}{4.5cm}{\centering \textbf{Evaluation (long premise filtered by BM25 retrieval)}}&\phantom{abc}  &\multicolumn{3}{@{}c}{\multirow{3}{2.2cm}{\centering \textbf{[Entail hypotheses as queries]}}}&  &\multicolumn{3}{@{}c}{\multirow{3}{2.2cm}{\centering \textbf{[Contradict hypotheses as queries]}}} \\
			&  &  &  &  &  &  &  &\\
			&  &  &  &  &  &  &  &\\
			\cmidrule{1-1} \cmidrule{3-5} \cmidrule{7-9}
			\centering \textbf{Recall@k}&&\textbf{1}&\textbf{10}&\textbf{100}&&\textbf{1}&\textbf{10}&\textbf{100}\\
			\midrule
\centering albert-xxlarge-v2\_anli-LawngNLI-retrieval&&0.138+/-0.018&0.321+/-0.019&0.505+/-0.003&&0.1+/-0.016&0.271+/-0.019&0.486+/-0.003\\
\centering albert-xxlarge-v2\_anli&&0.09+/-0.016&0.262+/-0.02&0.505+/-0.003&&0.07+/-0.014&0.174+/-0.019&0.486+/-0.003\\
\centering albert-xxlarge-v2\_ConTRoL-dataset&&0.034+/-0.011&0.144+/-0.018&0.505+/-0.003&&0.048+/-0.012&0.177+/-0.019&0.486+/-0.003\\
\centering albert-xxlarge-v2\_DocNLI&&0.067+/-0.014&0.226+/-0.02&0.505+/-0.003&&0.008+/-0.006&0.083+/-0.015&0.486+/-0.003\\
\centering all-mpnet-base-v2&&0.09+/-0.016&0.228+/-0.02&0.505+/-0.003&&0.076+/-0.015&0.219+/-0.019&0.486+/-0.003\\
\centering cross-encoder\_ms-marco-MiniLM-L-6-v2&&0.019+/-0.009&0.116+/-0.017&0.505+/-0.003&&0.013+/-0.008&0.123+/-0.017&0.486+/-0.003\\
\centering BM25&&0.016+/-0.008&0.061+/-0.014&0.505+/-0.003&&0.008+/-0.006&0.052+/-0.013&0.486+/-0.003\\
\midrule
			\centering N &  \multicolumn{8}{c}{1322} \\
			\bottomrule
		\end{tabular}
	}
	\caption{\label{tab:table35}
		Recall@k of model panel, when re-ranking the bi-encoder top 100 (ranked by all-distilroberta-v1 (\citealp{SDrCW19}; \citealp{LiuOtGoDuJoChLeLeZeSt19}) prepended with BM25 (\citealp{RobertsonZa09}) filtering from Appendix Table~\ref{tab:table34}) for implication-based retrieval. The error provided is the larger of the two deviations of the Clopper-Pearson (\citealp{ClopperPe34}) exact binomial 95\% confidence bounds from the point estimate.
	}
\end{table*}

We proceed in three steps: cheap retrieval, bi-encoder ranking, and cross-encoder re-ranking.

\noindent \textbf{(1) Building our test set, using cheap retrieval across all candidate cases:} Our retrieval test set comprises the LawngNLI test set's Entail and Contradict long premises (majority opinions, here before dataset processing) as positive examples, pooling each with 999 other majority opinions from the same state (or the federal level) selected by highest all-mpnet-base-v2 (\citealp{STQLL20}) embedding dot-product similarity with the hypothesis as negative examples. An effective retrieval system would rank the premise case highly as correct, and we can compare models on Recall@k. So each non-Neutral LawngNLI test set hypothesis is a query paired with 1000 candidate documents.

This step corresponds to the initial retrieval step in a standard retrieve-and-re-rank approach to the retrieval problem. We control for this first step by uniformly applying this method and bracket a search for improvements. This allows our comparison to focus on which models rank target cases most highly in steps 2 and 3, with the aspiration of rankings high enough to be within users' reach with minimal skimming beyond the top result. 

\noindent \textbf{(2) Zero-shot bi-encoder ranking of retrieved top 1000:} Our zero-shot panel comprises a BM25 (\citealp{RobertsonZa09}) baseline and four Sentence-Transformers (\citealp{ReimersGu21}) lightweight bi-encoders:  msmarco-distilroberta-base-v2, nli-distilroberta-base-v2, all-distilroberta-v1, and all-mpnet-base-v2 (\citealp{SDrCW19}; \citealp{LiuOtGoDuJoChLeLeZeSt19}; \citealp{BCCDGLMMMN16}; \citealp{STQLL20}). The first three were chosen for comparable setups besides their training task, while the last is a top model across semantic search evaluations.\footnote{\url{https://www.sbert.net/docs/pretrained_models.html\#sentence-embedding-models}. No all-distilroberta-v2 was available on HuggingFace (\citealp{WCDrSDMCFDS20}).} They are respectively pretrained for actual user queries from the Bing search engine, NLI, and a combined dataset of over 1 billion pairs of related sentences covering many tasks. As in Section~\ref{sec:generalization}, we also evaluate each model version while prepending with a module that filters each candidate document to the 5 paragraphs with highest BM25 similarity to the query.

Appendix Table~\ref{tab:table34} shows that leading, lightweight bi-encoders and BM25 (\citealp{RobertsonZa09}) can rank the target case reasonably well zero shot while processing a large number of candidate documents. Prepending with BM25 filtering improves performance. Since Entail and Contradict hypotheses are twinned, it is not so surprising that performance is similar between those two subsets. Still, these retrieval models seem capable of recognizing long-context inference (as a subtype of relevance) close to equally between entailed and contradicted queries. While Recall@10 of the target premise case reaches about 0.3, Recall@100 still only reaches about 0.5 (versus expected value of 0.1), showing that leading retrieval systems would miss important documents for legal retrieval zero shot.

\noindent \textbf{(3) NLI cross-encoder re-ranking of ranked top 100:} We re-rank our top Recall@100 rankings, achieved by all-distilroberta-v1 (\citealp{SDrCW19}; \citealp{LiuOtGoDuJoChLeLeZeSt19}) prepended with BM25 (\citealp{RobertsonZa09}) filtering. Re-ranking is by (Entail or Contradict) label probability. All models are prepended with BM25 filtering, which improved performance in Appendix Table~\ref{tab:table34}.

Our model panel starts with albert-xxlarge-v2 (\citealp{LCGGSS19}) fine-tuned on each of our three included previous NLI datasets (ANLI (\citealp{NWDBWK20}), ConTRoL (\citealp{LCLZ21}), and DocNLI (\citealp{YinRaXi21})). We also add cross-encoder/ms-marco-MiniLM-L-6-v2 (\citealp{WWDBYZ20}; \citealp{ReimersGu21}; \citealp{BCCDGLMMMN16}) as an additional zero-shot baseline. Among our dense baselines, fine-tuning on ANLI provides the top re-ranking performance.  We then fine-tune the ANLI  model on an adjusted retrieval version of LawngNLI.\footnote{Here, premises are without prepending and without citation spans removed. And models are instead fine-tuned on LawngNLI's Entail or Contradict examples, together with its Neutral hypotheses instead paired with a random incorrect all-mpnet-base-v2 (\citealp{STQLL20}) top 100 candidate in order to align more closely with re-ranking task negatives.} We use the same setup as for our NLI intermediate fine-tuning on ANLI except with a learning rate of 1e-6.

Table~\ref{tab:table35} presents our comparison. Zero shot, re-ranking with these dense models underperforms BM25 (\citealp{RobertsonZa09}) on Recall@1 and Recall@10 (compare Appendix Table~\ref{tab:table34}). However, fine-tuning on our adjusted version of LawngNLI draws re-ranking performance toward BM25. Refining our approach to fine-tuning using LawngNLI may yield future improvements.

\section{Related Work}
\label{sec:relatedwork}

\noindent \textbf{Legal retrieval:} \citet{LockeZu22} note that the most used commercial systems have not publicly released specific algorithms. Still, some providers have shared core feature types or modules. The retrieval module for a recent QA system (by authors including LexisNexis researchers) compares candidates by sparse (BM25-based) and dense representations (\citealp{KPMSSCCS21}). WestSearch Plus categorizes questions into frames and selects passages using QA-pair classifiers (\citealp{MSMTPC19}). Modern commercial systems' features include text, citation networks, annotations, and query logs.\footnote{\url{https://blog.law.cornell.edu/voxpop/2013/03/28/next-generation-legal-search-its-already-here/}} Published retrieval systems such as \citet{KPMSSCCS21} (by authors including LexisNexis researchers) and \citet{TangCl21} do leverage Transformer embedding similarity. To our knowledge, our application of NLI-based retrieval to legal case research is novel.

\noindent \textbf{Long-context NLI benchmarks:} SCROLLS (\citealp{SSIEYHGXGB22}) analyzes long-context datasets across multiple tasks (ContractNLI (\citealp{KoreedaMa21}) is the included NLI dataset). \citet{SCBFM22} investigated retrieval and aggregation methods to scale up their sentence-level NLI model and baselines for DocNLI (\citealp{YinRaXi21}) and ContractNLI (\citealp{KoreedaMa21}). Without premise multigranularity, however, these long-context experiments do not evaluate models' generalization to long contexts while holding constant the target domain and examples (e.g., no model-external NLI labels for evaluation data's evidence spans).

ContractNLI (\citealp{KoreedaMa21}) constructed an NLI dataset with similarly long premises, though containing a substantially smaller number of examples (607 contracts as premises, each paired with 17 shared hypotheses) in the quite linguistically different contracts domain. They study evidence identification and context segmentation within their premises, as compared to (in the present paper) benchmarking models on multigranular premises. LawngNLI contains over a hundred thousand distinct hypotheses, providing models with domainwide supervision.

\noindent \textbf{Other legal NLP:} AutoLAW and CaseHOLD (\citealp{Mahari21}; \citealp{ZGARHH21}) construct datasets for a distinct task of predicting holdings from other cases that support the arguments in the nearby context in the \textit{citing} case. These holdings exhibit an argument support relation with respect to their surrounding context, as opposed to necessarily any NLI relation. Recently, \citet{SLYDSD22} introduced a multi-document summarization dataset for documents around civil rights cases, with multiple granularities of summaries. The legal tasks closest to ours are from the annual COLIEE workshop.\footnote{\url{https://sites.ualberta.ca/~rabelo/COLIEE2021/}} However, these tasks do not fully map to three-label NLI. For, e.g., relevant 2021 Tasks 2 and 4, their training corpora (in the hundreds of examples) are ballpark 1000 times smaller than usual single-sentence benchmarks, making supervised learning alone insufficient for reliably training models to generalize (\citealp{HMCHBVS20}; \citealp{RGyKKYS21}; \citealp{KRGy21}; \citealp{SCMHVBH21}).

\section{Conclusion and Future Work}
\label{sec:conclusion}

This work presents LawngNLI, a new NLI benchmark with multigranular long premises, each containing a shorter version. Experiments demonstrate some use cases. First, we show that leading NLI models fall substantially short of competitive performance when generalizing to LawngNLI with its long premises, even after fine-tuning using existing NLI benchmarks and/or LawngNLI with short premises (with the same domain and examples as the long-premise evaluation). Unconfounded by domain shift, these results show the need for a large-scale long-premise dataset like ours at fine-tuning time.

Second, we show that leading lightweight retrieval models can reasonably handle implication-based retrieval on LawngNLI zero shot with both entailed and contradicted arguments as queries. We then compare re-ranking by lexical overlap and models fine-tuned using a modified LawngNLI or several previous NLI datasets. Multiple other aspects of LawngNLI are left for future study.

\section*{Limitations}

LawngNLI contains automatic labels, derived from the construction process. Its Entail labels are effectively annotated by judges, who wrote Entail hypotheses as parentheticals asserted by the cited premise. Neutral and Contradict examples are derived from Entail examples by, respectively, re-pairing with a different non-adjacent premise in the citation network and by adding or removing pivotal negation (Appendix Section~\ref{sec:datasetconstruction}). These steps could introduce some error rate, which we validate by human assessment (Section~\ref{sec:labels}). And using a human-validated subset, Section~\ref{sec:goldlabels} evinces that our conclusions from our generalization experiment likely do not arise from such differences between automatic and gold labels.

Both experiments test standard approaches when applied to distinct challenges: first, short-premise models on long-premise NLI and, second, semantic search models on implication-based retrieval. For the experiment on generalization to long contexts, we demonstrate that these standard approaches do not always suffice, but only in one (albeit important) counterexample domain: law. We have not established if these shortcomings extend to other domains more broadly.

\section*{Ethics Statement}
\label{sec:ethics}

Considerations for general NLI have been explored elsewhere (e.g., for gender bias by \citet{SharmaDeSi21}). We discuss some considerations for the legal aspect. On the benefit side, NLI is a principal cognitive task in law, so progress here also stands to benefit the legal community: Building court cases and advising clients essentially is arguing for and against different natural language inferences from legal texts and facts. Implications may not be directly stated in the text or annotations (e.g., those at a different level of specificity or requiring compositional reasoning). Instead, holdings and rules inferable from case text must be extracted through costly human annotation and curation. All around the legal system, the pay grade and spare bandwidth of legal counsel is frequently starkly imbalanced between parties with adversarial interests: whether people in the courtroom or settlement conference, consumers or companies in a negotiation boardroom, or in everyday society where behavior is shaped by prospects of legal action. Anything that makes legal research and thus legal counsel cheaper, including more lightweight or task-tailored case retrieval systems, can contribute toward fairer access to legal representation and justice regardless of financial means. Models that perform well on LawngNLI's retrieval setup could crosswalk between cases as premises and implications as hypotheses, performing implication-based retrieval automatically. We describe the state of legal retrieval systems (including limited public information about leading proprietary commercial algorithms) relative to an NLI-based approach in Section~\ref{sec:relatedwork}. Legal services overall comprise about 1.3\% of U.S. GDP.\footnote{\url{https://fred.stlouisfed.org/graph/?g=PfxD}} The legal research industry's annual revenue meanwhile is in the multiple billions of dollars.\footnote{As of 2020: e.g., \citet{Reuters}.} And the full societal cost of suboptimal case retrieval should include the time and resources expended by human legal researchers in the loop (paralegals and lawyers) in unnecessary iterating with any suboptimal retrieval systems. Indeed, junior lawyers (less than 10 years of experience) spend almost a third (28\%) of their working time on case research (\citealp{Poje14}).\footnote{This percentage that falls to 16\% for senior lawyers who are likely to take on a more managerial role.}

On the risk side, while prospective human reliance for decision making on erroneous model predictions is an ever-present consideration in NLP, we do not view this as a practical risk for LawngNLI. Everyday people can turn to numerous simple articles online summarizing the law, without digging into complex case retrieval and jurisprudence. And regarding advising others, lawyers bound by professional duties are exclusively authorized to practice law in the U.S. and around the world.\footnote{\url{https://www.ibanet.org/MediaHandler?id=199b20ec-b7ab-4ef4-99c4-cd45c7b6371b}} Nothing can even be done just knowing the most relevant cases or implications; they must be synthesized by human judgment into an argument sound enough to pass the muster of judges and juries. In other words, legal NLI models are in no way lawyers. Instead, they can work as screening tools for practitioners who then must apply their own judgment to make the results useful. In this way, legal NLI models could help save the resources of lawyers and clients and help improve the quality of legal representation.

\section*{Acknowledgements}

This research is based upon work supported in part by the Oﬃce of the Director of National Intelligence (ODNI), Intelligence Advanced Research Projects Activity (IARPA), via IARPA Contract No. 2019-19051600006 under the BETTER Program. The views and conclusions contained herein are those of the authors and should not be interpreted as necessarily representing the oﬃcial policies, either expressed or implied, of ODNI, IARPA, the Department of Defense, or the U.S. Government. The U.S. Government is authorized to reproduce and distribute reprints for governmental purposes notwithstanding any copyright annotation therein. This research is supported by a Focused Award from Google.

\bibliography{ccg, cited_10s, cited_00s, cited_95s, cited_90s, cited_pre90s, new}
\bibliographystyle{acl_natbib}

\appendix

\section{Appendix}

\subsection{Dataset Construction Procedure}
\label{sec:datasetconstruction}

\subsubsection{Extraction from Caselaw Access Project}

LawngNLI is constructed starting with all xml case files from the April 21, 2021 bulk export from the Caselaw Access Project (\citealp{PresidentUn18}). The word count of the full original corpus before processing at about 12 billion\footnote{\url{https://case.law/docs/site_features/trends}} is around three times that of English Wikipedia\footnote{About 4 billion as of December 1, 2021: \url{https://web.archive.org/web/20211201013917/https://en.wikipedia.org/wiki/Special:Statistics}}, though for our premises we limit to only the majority opinions.

Entail examples are pairs of citation parentheticals (hypotheses) and excerpts of majority opinions from cited cases with resolvable pincites (premises), extracted from case files using Eyecite (\citealp{CushmanDaLi21}). Where Eyecite associates multiple consecutive citations resolving to the same case with the same citation parenthetical, only the first citation and its pincite, if any, is paired with the parenthetical and included as an example. In this paper, we only include examples from citations including a resolvable pincite (e.g., does not contain letters).

The short version of the premise consists of the resolvable cited pages within the cited case's majority opinion, while the long version of the premise consists of the cited case's full majority opinion. 

\subsubsection{Initial Filters}

Examples are dropped or modified by simple ``accuracy'' filters. First, as an overbroad criterion to exclude examples where the (converted or unconverted) original Entail hypothesis was a parenthetical in a case that was later overturned, we drop all examples with hypotheses from cases where a later case shared the same party names in the same or reverse order.
	
Second, parentheticals with citations including a case history flag (e.g., ``acq.'',``aff'd'') are excluded.
	
Third, we drop examples with hypotheses that contain certain regex keywords (`quoting|en banc|omitted|mphasis|applying|citing |concur|dissent|majority|, in chambers|per curiam|Lexis|opinion| v. |§|¶|[0-9]') associated with parentheticals describing ``metadata'' about the cited case rather than its content.
	
Fourth, verbs ending with ``ing'' followed by ``that'' at the beginning of remaining hypotheses many times take a supporting stance toward the subsequent subordinate clause, so to adapt such hypotheses to be more similar to a standalone sentence, we remove such initial words and the subsequent ``that'' in hypotheses.
	
Finally, sentences are normalized with spaCy 3.1.1 (\citealp{MHHLeBPMSGOOAKRBFeHPTBMABTVBMGJH21}) to, e.g., process contractions.

\subsubsection{Identifying (Pivotal) Negation in Hypotheses}

Next the Entail examples are automatically labeled by whether their hypotheses contain (pivotal) negation or not, depending on whether the contradiction algorithm described in Appendix Section~\ref{sec:contradictionalgorithm} removes or adds negation, respectively. Pairs with hypotheses rejected for processing by our contradiction algorithm are dropped from the dataset.

Since the absence versus presence of such negation in the hypothesis results in contradictory truth values (and thus also flips the NLI label between Entail and `Contradict), such negation can be called ``pivotal.'' Negation is defined this way throughout the paper except in Table~\ref{tab:table1} when comparing to other datasets, since our contradiction algorithm might exhibit a different error rate on those datasets and confound the comparison. For this reason, greater than 50\% of LawngNLI's hypotheses contain negation in Table~\ref{tab:table1}, even though the dataset is constructed to contain 50\% (pivotal) negation hypotheses.  

\subsubsection{NLI Label Split}
\label{sec:nlilabelsplit}

Within examples from cases from each state (or federal) and pivotal negation or not, entail examples are randomly assigned to be 1/3 Entail, 1/3 converted to Neutral, and 1/3 converted to Contradict.

\subsubsection{Converting Entail Examples to Contradict Examples: Contradiction Algorithm}
\label{sec:contradictionalgorithm}

For examples labeled Contradict in Appendix Section~\ref{sec:nlilabelsplit}, we use our contradiction algorithm to add or remove pivotal negation\footnote{``Pivotal'' negation is negation the absence versus presence of which results in at least some contradictory truth values for the hypothesis, flipping its NLI label from Entail to Contradict.} from the hypothesis, toward aligning the NLI relation with the label.

Our contradiction algorithm builds on the negation algorithm outlined in Section 4.2 of \citet{BiluHeSl15}, which in their paper was annotated by majority vote to have generated an opposing claim with probability 0.79.\footnote{Hypotheses are parsed with the Berkeley Neural Parser 0.2.0 `benepar\_en3' with spaCy 3.1.1 `en\_core\_web\_lg' (\citealp{KitaevCaKl19}, \citealp{KitaevKl18}, \citealp{MHHLeBPMSGOOAKRBFeHPTBMABTVBMGJH21}). Verb tense is modified using NLTK 3.6.2 WordNet Lemmatizer and Pattern 3.6 conjugate function (\citealp{University10}; \citealp{BirdKlLo09}; \citealp{SmedtDa12}). We explored attempting to negate adjectives and verbs using the lexical negation dictionary compiled by \citet{SonMiMo16} but ultimately limited to just using direct negation.}

The algorithm chooses a random sentence for adding or removing negation and leaves the others unchanged. It finds a non-compound independent clause within the chosen sentence and then makes the first applicable change in the list below. If none of the changes' conditions apply, the hypothesis is rejected for processing by the algorithm. This includes rejecting hypotheses consisting of verb phrases not nested within independent clauses; since these are rarely found in negated form in the original dataset, including them would leave an artifact of this contradiction algorithm. So for these hypotheses, we prioritize balance across labels over coverage of candidate examples.

\begin{enumerate}
	\item If there are any contradictable indefinite pronouns in the first highest-level noun phrase, the first one is changed to a contradictory pronoun (e.g., ``some'' to ``none'' or ``neither'' to ``either'').
	\item If there are any verb phrases, the first highest-level verb phrase is contradicted using a modified version (e.g., also reversing negation by removing ``do''/``does''/``did''+``not'') of the negation algorithm from \citet{BiluHeSl15} mentioned above.
	\item If there are any adjective phrases, the first [`no',`not',`never'] is removed from or else a `not' is added to the first highest-level adjective phrase or past participle.
\end{enumerate}

\subsubsection{Filtering}

Now we apply simple ``difficulty'' filters: examples with hypotheses containing quotation marks or fewer than four words or with at least 50\% bigram overlap with their premise are dropped.

\subsubsection{Converting Entail Examples to Neutral Examples: Neutralization Algorithm}

For examples labeled Neutral in Appendix Section~\ref{sec:nlilabelsplit}, we use our neutralization algorithm to match the hypothesis with a different premise, toward aligning the NLI relation with the label. To balance attrition, the neutralization algorithm is applied to all examples regardless of NLI label, but only the hypotheses from Neutral examples are actually re-paired with the assigned premise.

The candidates for matching with each hypothesis are the premises from all examples that are from cases in the same state as the original premise (or from a federal case if the original premise is from a federal case). Excluded from candidacy are premises from cases citing or cited by the case containing the original hypothesis.

A hypothesis is paired with a candidate premise as follows. The short version of the premise is used for this step.

First, the top 30 (dot-product) nearest neighbors of the hypothesis among the candidates are retrieved using FAISS (\citealp{JohnsonDoJe19})\footnote{\url{https://github.com/facebookresearch/faiss}} on msmarco-distilbert-base-tas-b embeddings (\citealp{HLYLH21})\footnote{\url{https://huggingface.co/sebastian-hofstaetter/distilbert-dot-tas\_b-b256-msmarco}. \url{https://www.sbert.net/docs/pretrained-models/msmarco-v3.html} shows retrieval using dot-product similarity on this model's embeddings to perform best among several models on TREC-DL 2019 (\citealp{CMYCV20}) and the MS Marco Passage Retrieval dataset (\citealp{BCCDGLMMMN16}).} via Sentence-Transformers (\url{https://github.com/UKPLab/sentence-transformers}, \citealp{ReimersGu21}).

Second, candidate premises with which a hypothesis has at least 50\% bigram overlap are dropped. This step preserves the filter applied earlier to all examples through the re-pairing for the Neutral examples.

Finally, Neutral hypotheses only are paired with their remaining candidate premise with respect to which it has the highest BM25 (\citealp{RobertsonZa09}) score via Gensim 3.8.3 (\citealp{RehurekSo10}). For hypotheses of all labels, if no candidate premises remain, their example is dropped.

\subsubsection{Balancing}

We split the dataset into ``analysis''/non-``analysis'' subsets by the inclusion criterion for this paper's experimental evaluation (Section~\ref{sec:experimentalevaluation}): whether the sequence length of an example's long premise is at most 4096 tokens, via a RoBERTa (\citealp{LiuOtGoDuJoChLeLeZeSt19}) tokenizer.

Within each of the ``analysis''/non-``analysis'' subsets, the dataset is then downsampled by randomly sampling each of the three label-plus-negation groups closed under the contradiction operation (Entail+negation plus Contradict+non-negation; Contradict+negation plus Entail+non-negation; Neutral+negation plus Neutral+non-negation) down to the minimum of their example counts. A 90/5/5 train/val/test split is stratified by ``analysis''/non-``analysis'' subset and these groups.

Each example is then complemented with its contradictory twin: the same premise paired with the hypothesis modified by adding or removing pivotal negation (so applying the contradiction algorithm). Neutral labels are unchanged from the original example, while Entail and Contradict labels are flipped. This twinning balances the dataset within the ``analysis''/non-``analysis'' subsets on NLI label by pivotal negation versus not.

\subsubsection{Citation Removal Algorithm and Prepending}

Our algorithm here attempts to remove as many in-line citations from premises as it can so that the premises are more customary English-language texts. The processed premises are studied in this paper. But the dataset obtainable from code to be released will include the pre-processing premises as well for future study. Finally, we copy and prepend at the beginning of the long premises the minimum number of paragraphs from the end that contain 512 tokens, to limit models from relying on cues for the NLI label near the start.

\subsection{Implementation Details}
\label{sec:implementation}

External code is from GitHub repositories, with repository forking permitted under contemporaneous GitHub's Terms of Service. External models are from HuggingFace Transformers (\citealp{WCDrSDMCFDS20}; contemporaneously governed by an Apache License 2.0 permitting modification, distribution, etc.) or from GitHub repositories. Cases from the Caselaw Access Project (\citealp{PresidentUn18}) are used to construct our datasets. Any dataset sharing will comply with Caselaw Access Project (\citealp{PresidentUn18}) terms of access or else any separate agreement with the licensor. In particular, if necessary to ensure this compliance, we will share code for constructing our datasets rather than the datasets themselves.

NVIDIA ~12GB TITAN Xp, ~11GB GeForce GTX 1080 Ti, ~11GB GeForce RTX 2080 Ti, ~24GB TITAN RTX GPUs, and NVIDIA ~48GB RTX A6000 GPUs were used for all fine-tuning.

\subsubsection{Evaluation 1}

For our intermediate fine-tuning, we adapt the code and largely follow the respective model hyperparameters and fine-tuning settings of the three existing NLI benchmarks. The settings that we modify rather than follow are: attention gradient checkpointing, GPU setup while not changing accumulated batch size, and maximum sequence length (given our long sequence lengths, we also train for 3 epochs instead of 5 on DocNLI (\citealp{YinRaXi21})). Maximum sequence lengths for intermediate fine-tuning are the lesser of the model maximum and 2048 (except for a maximum sequence length of 156 for pretrained short-sequence models fine-tuned on ANLI, consistent with \citet{NWDBWK20}\footnote{\url{https://github.com/facebookresearch/anli}}). 

After intermediate fine-tuning, the long-sequence models' maximum sequence lengths are increased to 4096 for further fine-tuning on LawngNLI. We adapt the code from \citet{XZCTFLS21}.\footnote{\url{https://github.com/mlpen/Nystromformer}} We adapted this code in order to allow compatibility with their suite of efficient Transformers, but ultimately we did not pretrain them and did not further explore including them after several (initialized with copied RoBERTa-base (\citealp{LiuOtGoDuJoChLeLeZeSt19}) embeddings) did not rise far above random accuracy for LawngNLI fine-tuning under some initial hyperparameters explored. This reflects little on these models since we did not pretrain them.

For fine-tuning on LawngNLI in our NLI experiments, we use a batch size of 32, learning rate of 1e-5, 4 epochs (2 epochs for DocNLI; see next paragraph), learning rate schedule adapted from \citet{XZCTFLS21} (Adam optimizer with $\beta$1 = 0.9, $\beta$2 = 0.999, L2 weight decay of 0.01, warm-up over first 10,000 steps, and linear decay), and half precision. We explored hyperparameters among those explored by RoBERTa (\citealp{LiuOtGoDuJoChLeLeZeSt19}) for GLUE (\citealp{WangSiMiHiLeBo18}), along with batch size 128 so that all of our models in Appendix Section~\ref{sec:pretrained} would start to converge during fine-tuning starting from their initial losses and accuracies.

To transfer learning from two-label DocNLI, the models intermediate-fine-tuned on DocNLI are further fine-tuned and evaluated on a two-label version of LawngNLI (where the Entail examples are duplicated and then (Entail, Neutral and Contradict) labels are mapped to (Entail, Not Entail)). This construction balances the two-label version between (Entail, Not Entail). For further fine-tuning these models on LawngNLI, the number of epochs is then halved. This is equivalent to splitting the Neutral and Contradict examples (now labeled Not Entail) in the original three-label dataset in half across pairs of consecutive original epochs (1 and 2, 3 and 4, and so on) so that the fine-tuning example count is 2/3 of the original dataset times the original number of epochs, except that example shuffling also pools examples between these consecutive original epochs.

\subsubsection{Existing NLI Datasets}
\label{sec:existingnlidatasets}

For models in Appendix Section~\ref{sec:pretrained} with fine-tuned checkpoints provided at \url{https://github.com/facebookresearch/anli} (ALBERT-xxlarge-v2 (\citealp{LCGGSS19}), BART-large (\citealp{LLGGMLSZ20}), and RoBERTa-large (\citealp{LiuOtGoDuJoChLeLeZeSt19})), we used these model checkpoints. Otherwise we fine-tuned the models, aiming to replicate the original hyperparameters.

\subsection{Procedure for Human Assessment}
\label{sec:humanassessment}

Human assessment was limited to Amazon Mechanical Turk Master Workers based in the U.S. 

Assessed accuracy of examples with long premises is lower than for with short premises, even though the former arguably should have a higher accuracy against the ground truth: they are a superset of the information in the short premise, thereby providing additional context while being written to be internally consistent. It may be then that the human-assessed error rates for the automatic labels are themselves imperfect against the ground truth, especially for more difficult examples.

Human assessment proceeded as follows:

\begin{itemize}
	\item Examples were each reviewed by two workers in batches of 28 examples, which were drawn from a first and then second set of 504 examples with sequence length at most 4096. Each set consists of a stratified random sample of test examples. The stratification is as follows: First, balance over the Cartesian product of the automatic label and pivotal negation versus not. Then half using the short premise and half using the long premise.
	\item Workers provided NLI labels for batches effectively without a time limit (batches due 1 week after assignment). Batches were issued until there were 300 non-screening examples with their two worker labels in agreement. The accuracy of these examples' automatic labels was then evaluated against those agreed labels (as gold).
	\item Workers were advised that they were providing NLI labels to be used in an academic analysis evaluating a new dataset. 
	\item A co-author provided NLI labels for a predetermined random sample as ``screening'' examples, \textit{which did not enter the dataset with human-assessed gold labels or the error rate calculations and so do not directly impact them}. They were scattered throughout and not separately identified to workers. Performance formulae using the screening examples \textit{only} were used to calculate worker bonuses and to exclude (ultimately two) workers who appeared to be guessing frequently.	
	\item Workers were paid above the U.S. federal minimum wage on ``reasonable'' (as opposed to actual) time spent: 2 hours per batch, but workers may have spent more or less time on any batch up to 1 week. In addition, a performance bonus was provided for each label deemed correct on a screening example.
	\item Worker screening was as follows: 
	\begin{itemize}
		\item First, workers needed to qualify by answering at least 4 examples correct (credit was sometimes given for an incorrect label with defensible reasoning) on an initial pre-screen of six screening examples within a half hour. Several batches not meeting the minimum performance described in the instructions (which was itself below the qualification threshold) were rejected.
		\item Because NLI is multiple choice, there is a risk that the initial screening may be insufficient or that workers may not consider examples thoroughly in selecting options (or even guess somewhat randomly). Though we saw evidence directly in the gold dataset, we included screening examples in the ongoing batches. We excluded two workers' examples for falling below a threshold.
		\begin{itemize}
			\item Each batch contains 3 screening examples and 25 non-screening examples.
			\item Labels on screening examples were selected by a co-author. Screening examples were not included in the 300 examples in the gold dataset.
			\item Workers could continue completing the batches of 28 unless at a time of audit their cumulative accuracy on screening examples fell below 50\% (after at least 5 screening examples). If their cumulative accuracy fell below this threshold, they were still paid for all completed batches but the examples they labeled were not included in the gold dataset.
		\end{itemize}		
		\item Workers provided labels via a six-option scale: 'definitely entail', 'probably entail', 'definitely neutral', 'probably neutral', 'definitely contradict', 'probably contradict'.
		\item For examples that workers labeled as entail or contradict, they also copied and pasted a portion of the premise relevant to determining the label they chose.
		\item We temporarily experimented with having a different version of the dataset assessed, but no workers labeled the same examples in that experiment and the current assessment set.
	\end{itemize}
\end{itemize}

\subsubsection{Interface for Main NLI Task}

See Appendix Table~\ref{tab:figure1}.

\begin{table*}
	\centering
	\includegraphics[scale=0.36]{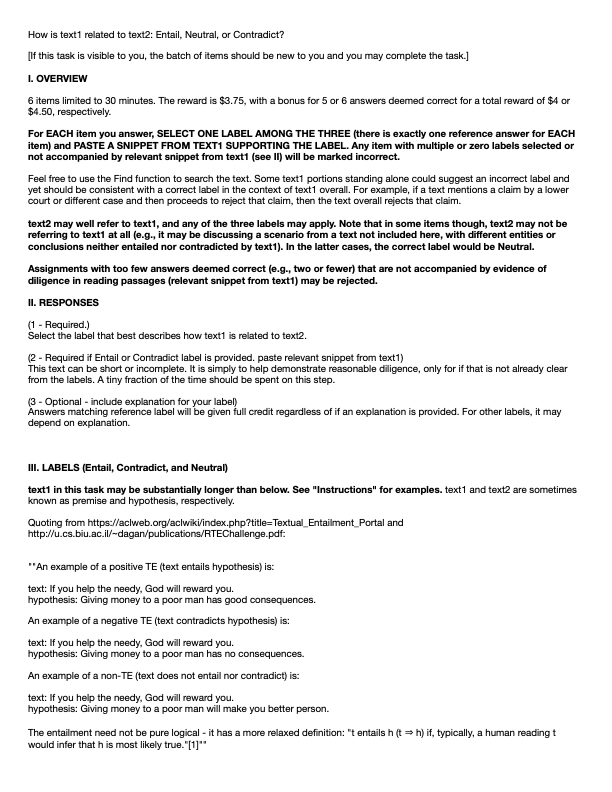}\includegraphics[scale=0.36]{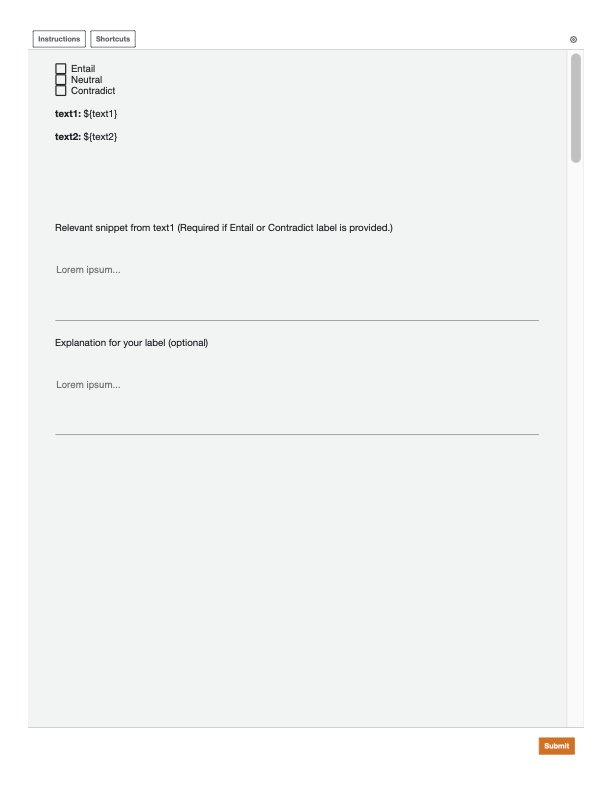}
	\caption{\label{tab:figure1}
		Interface for the main NLI task.
	}
\end{table*}

\subsubsection{Interface for Pre-Screen Task}

See Appendix Table~\ref{tab:figure2}. An included illustrative example is also omitted here. Note that some earlier workers saw earlier versions.

\begin{table*}
	\centering
	\includegraphics[scale=0.36]{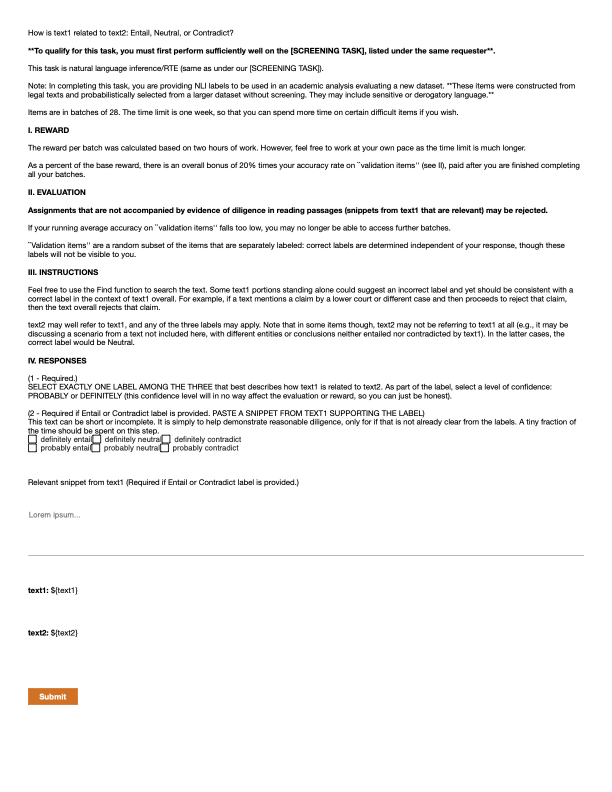}
	\caption{\label{tab:figure2}
		Interface for the pre-screen task. An included illustrative example is also omitted here. Note that some earlier workers saw earlier versions.
	}
\end{table*}

\subsection{Evaluation Panel: List of State-of-the-Art Pretrained Models}
\label{sec:pretrained}

\begin{itemize}
	\setlength{\itemsep}{0pt}
	\setlength{\parskip}{0pt}
	\setlength{\parsep}{0pt}
	\setlength{\topsep}{0pt}
	\setlength{\partopsep}{0pt}
	\item Longformer-base (\citealp{BeltagyPeCo20})
	\item BigBird-RoBERTa-base (\citealp{ZGDAAOPRWY})
	\item ALBERT-xxlarge-v2 (\citealp{LCGGSS19}). It ranked highest on ANLI test A2 and A3\footnote{\url{https://paperswithcode.com/sota/natural-language-inference-on-anli-test}}. It also ranked highest besides T5 models and third overall on MNLI (\citealp{WilliamsNaBo18})\footnote{\url{https://paperswithcode.com/sota/natural-language-inference-on-multinli}}.
	\item BART-large (\citealp{LLGGMLSZ20}). It ranked first on ConTRoL (\citealp{LCLZ21}), after fine-tuning on ANLI (\citealp{NWDBWK20}).
	\item Custom Legal-BERT (\citealp{ZGARHH21}). Pretrained on the Caselaw Access Project (\citealp{PresidentUn18}) corpus.
	\item  LEGAL-BERT-base-uncased, also known as LEGAL-BERT-SC (\citealp{CFMAA20}). It is pretrained on legal text from fields such as legislation, cases, and contracts.
	\item RoBERTa-large (\citealp{LiuOtGoDuJoChLeLeZeSt19}). It performed the better out of two models (over Longformer (\citealp{BeltagyPeCo20})) on DocNLI (\citealp{YinRaXi21}) and ranked second on ANLI test A1\footnote{\url{https://paperswithcode.com/sota/natural-language-inference-on-anli-test}}.
\end{itemize}

\subsection{Appendix Tables}

\begin{table*}
	\centering
	{\small
		\begin{tabular}{p{2cm}|p{13cm}}\toprule
			\multicolumn{2}{c}{\textbf{Sample twin Entail/Contradict examples with same premise from LawngNLI}}  \\
			\midrule
			\multirow{2}{1.9cm}{Twin hypotheses with same premise, from ``analysis'' subset} & \tabitem \textit{Contradict:} city acted affirmatively to create or increase risk of harm on city street by ignoring residents' requests to reduce speed limit or by taking down residents' signs indicating drivers should adhere to a lower speed limit \\
			& \tabitem \textit{Entail:} city did not act affirmatively to create or increase risk of harm on city street by ignoring residents' requests to reduce speed limit or by taking down residents' signs indicating drivers should adhere to a lower speed limit \\
			\midrule
			\multirow{2}{1.9cm}{Some additional hypotheses with same premise} & \tabitem \textit{Entail:} failing to enforce or lower the speed limit on a residential street ``did not create a `special danger' to a discrete class of individuals..[ed.: excerpted]..as opposed to a general traffic risk to pedestrians and other automobiles'' \\
			& \tabitem \textit{Contradict:} traffic laws and enforcement practices did not pose ``a general traffic risk to pedestrians and other automobiles'' \\
			\midrule
			Relevant excerpts of shared premise & \tabitem [ed.: Plaintiffs] ...submit that the City of Fort Thomas..violated their son's substantive due process rights by failing to act upon their request (and the requests of others) to lower the speed limit on the street..The police also removed signs posted by residents indicating that drivers should adhere to a 15 mile-per-hour speed limit..\\
			& \tabitem [ed.: Plaintiffs] ...alleged that the City's failure to maintain safe conditions on Garrison Avenue violated their son's substantive due process rights..established a ``state-created danger'' under DeShaney..\\
			& \tabitem ...DeShaney's holding..precludes [ed.: Plaintiffs'] argument that the Due Process Clause constitutionalizes a locality's choices about what speed limit to adopt for a given street or how to enforce that speed limit..\\
			& \tabitem There are two exceptions to the DeSha-ney rule..Under the second exception..a plaintiff may bring a substantive due process claim by establishing (1) an affirmative act by the State that either created or increased the risk that the plaintiff would be exposed to private acts of violence..\\
			& \tabitem [ed.: Plaintiffs] fail to satisfy any of the three requirements for establishing our circuit's ``state-created danger'' exception to DeShaney. First, the creation of a street and the management of traffic conditions on that street are too attenuated and indirect to count as an ``affirmative act''..\\
			\midrule
			Distractor excerpts of same premise & \tabitem ...After all, the City was told about the risks of not lowering the speed limit to 15 miles per hour (more accidents); it intentionally chose not to heed this warning (taking on the risk of more accidents); and the alleged risk came to pass when..was killed (an accident)..\\
			& \tabitem ...For in one sense, it could be said that all governing bodies act with deliberate indifference when they consider and reject a traffic-safety proposal of this sort that comes with known risks..\\
			\bottomrule
		\end{tabular}
	}
	\caption{\label{tab:sample1a}
		Sample twin Entail/Contradict examples with same premise from LawngNLI, also in the ``analysis'' subset analyzed in our experiments (Section~\ref{sec:experimentalevaluation}): sequence length of long premise at most 4096. Each hypothesis pairs with the excerpted premise in a separate example. For those specific ``additional hypotheses'' above, the examples containing them are in unfiltered-LawngNLI2 (see GitHub link in first footnote) but not LawngNLI, the core dataset studied in this paper. See also Table~\ref{tab:sample1}.}
\end{table*}

\begin{table*}
	\centering
	{\small
		\begin{tabular}{p{3cm}|p{12cm}}\toprule
			\multicolumn{2}{c}{\textbf{Sample twin Neutral examples with same premise from LawngNLI}} \\
			\midrule
			\multirow{3}{3cm}{Twin hypotheses with same premise, from ``analysis'' subset} & \tabitem \textit{Neutral:} a parade permit requirement did not violate the First Amendment \\
			& \tabitem \textit{Neutral:} a parade permit requirement violated the First Amendment \\
			& \\
			\midrule
			\multirow{2}{3cm}{Distractor excerpts of same premise} & \tabitem ...Section 13k prohibits two distinct activities: it is unlawful either ``to parade, stand, or move in processions or assemblages in the Supreme Court Building or grounds,''..\\
			& \tabitem ...we shall address only whether the proscriptions of 13k are constitutional as applied to the public sidewalks..\\
			\bottomrule
		\end{tabular}
	}
	\caption{\label{tab:sample2a}
		Sample twin Neutral examples from LawngNLI, but not in the ``analysis'' subset analyzed in our experiments (Section~\ref{sec:experimentalevaluation}): sequence length of long premise at most 4096. Each hypothesis pairs with the excerpted premise in a separate example. See also Table~\ref{tab:sample1}.}
\end{table*}

\begin{table*}
	\centering
	{\scriptsize
		\begin{tabular}{p{2cm}|p{13cm}}\toprule
			\multirow{2}{1.9cm}{Example 1 (human Neutral, automatic Entail)} & \tabitem \textit{Relevant premise excerpts:} Mebane’s sentence was clearly imposed by a court with jurisdiction, and his sentence is unambiguous. His sentence conforms to the statutory provisions regarding class A, B, and C felonies and the Habitual Criminal Act.\\
			& \tabitem \textit{Hypothesis:} claim that multiple sentences arose from single wrongful act and violated Double Jeopardy Clause does not establish that sentence is illegal \\
			\midrule
			\multirow{2}{1.9cm}{Example 2 (human Entail, automatic Contradict)} & \tabitem \textit{Relevant premise excerpts:} Moreover,(a)(6) states that “[a] report of an interdisciplinary team that contains the evaluation and signature of an acceptable medical source is also considered acceptable medical evidence,” while later in that section the statute designates nurse practitioners as an “other source.” 416.913(e)(3). While nowhere in the regulations is the term “interdisciplinary team” expressly defined, a plain reading of these sections taken together indicates that a nurse practitioner working in conjunction with a physician constitutes an acceptable medical source, while a nurse practitioner working on his or her own does not. \\
			& \tabitem \textit{Hypothesis:} non-medical source must not be integral to team, and the acceptable medical source must undersign her findings \\
			\midrule
			\multirow{2}{1.9cm}{Example 3 (human Contradict, automatic Neutral)} & \tabitem \textit{Relevant premise excerpts:} Where the BIA adopts the decision of the IJ, we review the IJ’s decision. The standard of review for factual findings made by the IJ is a deferential substantial evidence standard, and credibility findings will be upheld unless the evidence compels a contrary result. This deferential standard of review “precludes relief absent a conclusion that no reasonable factfinder could have reached the agency’s result.” Thangaraja v. Gonzales. \\
			& \tabitem \textit{Hypothesis:} the court may not make independent credibility determinations \\
			\bottomrule
		\end{tabular}
	}
	\caption{\label{tab:label}
		Several examples with differing automatic and gold labels.}
\end{table*}

\begin{table*}
	\centering
	{\scriptsize
		\begin{tabular}{@{}p{4.5cm}@{}|@{}p{0.1cm}@{}c@{}c@{}c@{}c@{}c@{}}\toprule
			\multirow{3}{4.5cm}{\centering \textbf{Fine-tuning and evaluation}}&&\multirow{3}{2.2cm}{\centering \textbf{Hypothesis only}}&\multirow{3}{2.2cm}{\centering \textbf{Long premise only (after BM25 filtering)}}&\multirow{3}{2.2cm}{\centering \textbf{Long premise only}}&\multirow{3}{2.2cm}{\centering \textbf{Short premise only (after BM25 filtering)}}&\multirow{3}{2.2cm}{\centering \textbf{Short premise only}}\\
			\centering &&&&&&\\
			\centering &&&&&&\\
			\midrule
			\multicolumn{7}{c}{[Entail/Neutral/Contradict. Chance=1/3]}\\
			\midrule
\centering albert-xxlarge-v2\_anli&&0.512+/-0.016&0.438+/-0.016&0.433+/-0.016&0.491+/-0.016&0.496+/-0.016\\
\centering albert-xxlarge-v2\_ConTRoL-dataset&&0.525+/-0.016&0.437+/-0.016&0.436+/-0.016&0.496+/-0.016&0.489+/-0.016\\
\centering albert-xxlarge-v2\_vanilla&&0.541+/-0.016&0.434+/-0.016&0.433+/-0.016&0.485+/-0.016&0.489+/-0.016\\
\centering &&&&&&\\
\centering facebook\_bart-large\_anli&&0.544+/-0.016&0.447+/-0.016&0.45+/-0.016&0.505+/-0.016&0.511+/-0.016\\
\centering facebook\_bart-large\_ConTRoL-dataset&&0.542+/-0.016&0.439+/-0.016&0.425+/-0.016&0.509+/-0.016&0.514+/-0.016\\
\centering facebook\_bart-large\_vanilla&&0.488+/-0.016&0.409+/-0.015&0.437+/-0.016&0.517+/-0.016&0.513+/-0.016\\
			\midrule
			\centering N &  \multicolumn{6}{c}{3966} \\
			\bottomrule
		\end{tabular}
	}
	\caption{\label{tab:table11}
		Accuracy of baselines on LawngNLI's ``analysis'' subset: Top 6 three-label models on both short and long premises from Appendix Table~\ref{tab:table32a}. The error provided is the larger deviation of the Clopper-Pearson (\citealp{ClopperPe34}) exact binomial 95\% confidence bounds.
	}
\end{table*}

\begin{table*}
	\centering
	{\scriptsize
		\begin{tabular}{@{}p{4.5cm}@{}|@{}p{0.1cm}@{}ccc@{}c@{}c@{}c@{}c@{}c@{}c@{}}\toprule
			\multirow{3}{4.5cm}{\centering \textbf{Needs long premises for fine-tuning}}&  &  &  &  &  &  &  &  &  &\\
			&  &  &  &  &  &  &  &  &  &\\
			&&\multicolumn{5}{c}{\textbf{No}}&&\multicolumn{3}{c}{\textbf{Yes}}\\
			\cmidrule{1-1} \cmidrule{3-7} \cmidrule{9-11}
			\multirow{3}{4.5cm}{\centering \textbf{Fine-tuning}}	 &
			\phantom{abc}&\multicolumn{3}{@{}c}{\multirow{3}{2.2cm}{\centering \textbf{Short premise}}}&
			\phantom{abc}&\multirow{3}{1.8cm}{\centering \textbf{BM25 retrieval on short premise}}&
			\phantom{abc}	 &\multirow{3}{1.8cm}{\centering \textbf{Long premise}}&
			\phantom{abc}&\multirow{3}{1.8cm}{\centering \textbf{Long premise filtered by BM25 retrieval}}\\
			&  &  &  &  &  &  &  &  &  &\\
			&  &  &  &  &  &  &  &  &  &\\
			\cmidrule{1-1} \cmidrule{3-5} \cmidrule{7-7} \cmidrule{9-9} \cmidrule{11-11}
			\multirow{4}{4.5cm}{\centering \textbf{Long premise filtered by BM25 retrieval at evaluation}} && \textbf{No} & \textbf{No} & \textbf{Yes} && \textbf{Yes} && \textbf{No} && \textbf{Yes}\\
			&  &  & \textbf{[512} & &  &  &  &  &  &\\
			&  &  & \textbf{[tokens]} & &  &  &  &  &  &\\
			&  & \textbf{(1)} & \textbf{(2)} & \textbf{(3)} &  & \textbf{(4)} &  & \textbf{(5)} &  & \textbf{(6)}\\
			\midrule
			\multicolumn{11}{c}{[Entail/Neutral/Contradict. Chance=1/3]}\\
			\midrule
			\centering albert-xxlarge-v2\_anli&&0.742+/-0.014&0.742+/-0.014&0.81+/-0.013&&0.817+/-0.012&&0.789+/-0.013&&0.853+/-0.011\\
			\centering albert-xxlarge-v2\_ConTRoL-dataset&&0.725+/-0.014&0.725+/-0.014&0.804+/-0.013&&0.806+/-0.013&&0.781+/-0.013&&0.851+/-0.011\\
			\centering albert-xxlarge-v2\_vanilla&&0.743+/-0.014&0.743+/-0.014&0.816+/-0.012&&\textbf{0.825+/-0.012}&&0.781+/-0.013&&0.852+/-0.011\\
			&&&&&&&&&&\\
			\centering allenai\_longformer-base-4096\_anli&&0.562+/-0.016&0.652+/-0.015&0.706+/-0.014&&0.707+/-0.014&&0.691+/-0.015&&0.784+/-0.013\\
			\centering allenai\_longformer-base-4096\_ConTRoL-dataset&&0.56+/-0.016&0.641+/-0.015&0.698+/-0.015&&0.691+/-0.015&&0.693+/-0.015&&0.781+/-0.013\\
			\centering allenai\_longformer-base-4096\_vanilla&&0.541+/-0.016&0.604+/-0.015&0.691+/-0.015&&0.701+/-0.015&&0.51+/-0.016&&0.791+/-0.013\\
			&&&&&&&&&&\\
			\centering facebook\_bart-large\_anli&&0.664+/-0.015&0.675+/-0.015&0.808+/-0.013&&0.809+/-0.013&&0.76+/-0.014&&0.868+/-0.011\\
			\centering facebook\_bart-large\_ConTRoL-dataset&&0.665+/-0.015&0.666+/-0.015&0.809+/-0.013&&0.814+/-0.012&&0.758+/-0.014&&0.868+/-0.011\\
			\centering facebook\_bart-large\_vanilla&&0.667+/-0.015&0.699+/-0.015&0.813+/-0.012&&0.807+/-0.013&&0.782+/-0.013&&\textbf{0.875+/-0.011}\\
			&&&&&&&&&&\\
			\centering google\_bigbird-roberta-base\_anli&&0.613+/-0.015&0.666+/-0.015&0.764+/-0.014&&0.774+/-0.013&&0.77+/-0.013&&0.829+/-0.012\\
			\centering google\_bigbird-roberta-base\_ConTRoL-dataset&&0.601+/-0.015&0.656+/-0.015&0.761+/-0.014&&0.769+/-0.013&&0.757+/-0.014&&0.819+/-0.012\\
			\centering google\_bigbird-roberta-base\_vanilla&&0.648+/-0.015&0.657+/-0.015&0.767+/-0.013&&0.777+/-0.013&&0.763+/-0.014&&0.827+/-0.012\\
			&&&&&&&&&&\\
			\centering nlpaueb\_legal-bert-base-uncased\_anli&&0.709+/-0.014&0.709+/-0.014&0.777+/-0.013&&0.791+/-0.013&&0.767+/-0.013&&0.811+/-0.013\\
			\centering nlpaueb\_legal-bert-base-uncased\_ConTRoL-dataset&&0.721+/-0.014&0.721+/-0.014&0.766+/-0.014&&0.787+/-0.013&&0.761+/-0.014&&0.816+/-0.012\\
			\centering nlpaueb\_legal-bert-base-uncased\_vanilla&&0.729+/-0.014&0.729+/-0.014&0.776+/-0.013&&0.788+/-0.013&&0.767+/-0.013&&0.813+/-0.012\\
			&&&&&&&&&&\\
			\centering roberta-large\_anli&&0.716+/-0.014&0.716+/-0.014&0.778+/-0.013&&0.799+/-0.013&&0.778+/-0.013&&0.842+/-0.012\\
			\centering roberta-large\_ConTRoL-dataset&&0.707+/-0.014&0.707+/-0.014&0.759+/-0.014&&0.789+/-0.013&&0.761+/-0.014&&0.838+/-0.012\\
			\centering roberta-large\_vanilla&&0.718+/-0.014&0.718+/-0.014&0.777+/-0.013&&0.801+/-0.013&&0.766+/-0.013&&0.843+/-0.012\\
			&&&&&&&&&&\\
			\centering zlucia\_custom-legalbert\_anli&&0.723+/-0.014&0.723+/-0.014&0.794+/-0.013&&0.805+/-0.013&&0.776+/-0.013&&0.824+/-0.012\\
			\centering zlucia\_custom-legalbert\_ConTRoL-dataset&&0.727+/-0.014&0.727+/-0.014&0.792+/-0.013&&0.802+/-0.013&&0.782+/-0.013&&0.824+/-0.012\\
			\centering zlucia\_custom-legalbert\_vanilla&&0.732+/-0.014&0.732+/-0.014&0.785+/-0.013&&0.798+/-0.013&&0.783+/-0.013&&0.832+/-0.012\\
			\midrule
			\centering Maximum of p-values versus (6) &&0&0&0&&0&&&&\\
			\midrule
			\centering N &  \multicolumn{10}{c}{3966} \\
			\bottomrule
		\end{tabular}
	}
	\caption{\label{tab:table32a}
		Accuracy on LawngNLI's ``analysis'' subset \textit{with long premises}: Full intermediate-fine-tuned model panel. Pretrained models are chosen for top performance on existing NLI benchmarks (see Appendix Section~\ref{sec:pretrained} for versions). The error provided is the larger deviation of the Clopper-Pearson (\citealp{ClopperPe34}) exact binomial 95\% confidence bounds. All p-values round to zero ($<$0.0005) from an exact binomial McNemar's (\citealp{McNemar47}) test for a statistically significant difference in accuracies between each version fine-tuning using short premises as inputs (1-4) and the best version fine-tuning using long premises as inputs (6). For (2), 512 tokens is the overall sequence limit.
	}
\end{table*}

\begin{table*}
	\centering
	{\small
		\begin{tabular}{@{}c|c@{}c@{}cccc@{}}\toprule
			\textbf{Evaluation}	 &\multicolumn{6}{c}{\textbf{Long Premise}}\\
			\midrule
			\textbf{Fine-tuning (on same)} && \textbf{No} && \multicolumn{3}{c}{\textbf{Yes}}\\
			\cmidrule{1-1} \cmidrule{3-3} \cmidrule{5-7}
			\textbf{Subset} && \textbf{All} && \textbf{All} & \multirow{3}{2cm}{\centering \textbf{Above-Median Long Premise Length}} & \multirow{3}{2cm}{\centering \textbf{Hypothesis Contains Negation}}\\
			&&&&&&\\
			&&&&&&\\
			\midrule
			\multicolumn{7}{c}{[Entail/Neutral/Contradict. Chance=1/3]}\\
			\midrule
			\centering albert-xxlarge-v2\_anli&&0.501+/-0.016&&0.789+/-0.013&0.738+/-0.02&0.789+/-0.019\\
			\centering albert-xxlarge-v2\_ConTRoL-dataset&&0.434+/-0.016&&0.781+/-0.013&0.732+/-0.02&0.787+/-0.019\\
			\centering albert-xxlarge-v2\_vanilla&&0.345+/-0.015&&0.781+/-0.013&0.733+/-0.02&0.784+/-0.019\\
			&&&&&&\\
			\centering allenai\_longformer-base-4096\_anli&&0.367+/-0.015&&0.691+/-0.015&0.652+/-0.022&0.695+/-0.021\\
			\centering allenai\_longformer-base-4096\_ConTRoL-dataset&&0.355+/-0.015&&0.693+/-0.015&0.664+/-0.022&0.692+/-0.021\\
			\centering allenai\_longformer-base-4096\_vanilla&&0.334+/-0.015&&0.51+/-0.016&0.462+/-0.023&0.511+/-0.022\\
			&&&&&&\\
			\centering facebook\_bart-large\_anli&&0.345+/-0.015&&0.76+/-0.014&0.695+/-0.021&0.767+/-0.019\\
			\centering facebook\_bart-large\_ConTRoL-dataset&&0.407+/-0.015&&0.758+/-0.014&0.699+/-0.021&0.764+/-0.019\\
			\centering facebook\_bart-large\_vanilla&&0.336+/-0.015&&0.782+/-0.013&0.723+/-0.021&0.789+/-0.019\\
			&&&&&&\\
			\centering google\_bigbird-roberta-base\_anli&&0.342+/-0.015&&0.77+/-0.013&0.74+/-0.02&0.773+/-0.019\\
			\centering google\_bigbird-roberta-base\_ConTRoL-dataset&&0.354+/-0.015&&0.757+/-0.014&0.735+/-0.02&0.757+/-0.019\\
			\centering google\_bigbird-roberta-base\_vanilla&&0.335+/-0.015&&0.763+/-0.014&0.747+/-0.02&0.761+/-0.019\\
			&&&&&&\\
			\centering nlpaueb\_legal-bert-base-uncased\_anli&&0.478+/-0.016&&0.767+/-0.013&0.725+/-0.02&0.775+/-0.019\\
			\centering nlpaueb\_legal-bert-base-uncased\_ConTRoL-dataset&&0.423+/-0.016&&0.761+/-0.014&0.726+/-0.02&0.768+/-0.019\\
			\centering nlpaueb\_legal-bert-base-uncased\_vanilla&&0.336+/-0.015&&0.767+/-0.013&0.725+/-0.02&0.769+/-0.019\\
			&&&&&&\\
			\centering roberta-large\_anli&&0.353+/-0.015&&0.778+/-0.013&0.727+/-0.02&0.781+/-0.019\\
			\centering roberta-large\_ConTRoL-dataset&&0.429+/-0.016&&0.761+/-0.014&0.717+/-0.021&0.763+/-0.019\\
			\centering roberta-large\_vanilla&&0.333+/-0.015&&0.766+/-0.013&0.712+/-0.021&0.772+/-0.019\\
			&&&&&&\\
			\centering zlucia\_custom-legalbert\_anli&&0.499+/-0.016&&0.776+/-0.013&0.736+/-0.02&0.779+/-0.019\\
			\centering zlucia\_custom-legalbert\_ConTRoL-dataset&&0.445+/-0.016&&0.782+/-0.013&0.753+/-0.02&0.783+/-0.019\\
			\centering zlucia\_custom-legalbert\_vanilla&&0.49+/-0.016&&0.783+/-0.013&0.753+/-0.02&0.787+/-0.019\\
			\midrule
			\centering N&&3966&&3966&1936&1983\\
			\midrule
			\multicolumn{7}{c}{[Entail/Not entail. Chance=1/2]}\\
			\midrule
			\centering albert-xxlarge-v2\_DocNLI&&0.422+/-0.013&&0.847+/-0.01&0.804+/-0.016&0.872+/-0.013\\
			\centering allenai\_longformer-base-4096\_DocNLI&&0.5+/-0.014&&0.777+/-0.011&0.76+/-0.017&0.814+/-0.015\\
			\centering facebook\_bart-large\_DocNLI&&0.497+/-0.014&&0.641+/-0.013&0.602+/-0.019&0.64+/-0.019\\
			\centering google\_bigbird-roberta-base\_DocNLI&&0.513+/-0.014&&0.817+/-0.011&0.798+/-0.016&0.843+/-0.014\\
			\centering nlpaueb\_legal-bert-base-uncased\_DocNLI&&0.48+/-0.014&&0.833+/-0.01&0.814+/-0.015&0.85+/-0.014\\
			\centering roberta-large\_DocNLI&&0.448+/-0.014&&0.843+/-0.01&0.82+/-0.015&0.877+/-0.013\\
			\centering zlucia\_custom-legalbert\_DocNLI&&0.557+/-0.014&&0.822+/-0.011&0.801+/-0.016&0.842+/-0.014\\
			\midrule
			\centering N&&5288&&5288&2637&2644\\
			\bottomrule
		\end{tabular}
	}
	\caption{\label{tab:table2a}
		Accuracy on LawngNLI's ``analysis'' subset: Full intermediate-fine-tuned model panel. If fine-tuned, fine-tuning on the LawngNLI subset is on premises with the same granularity as evaluation. The error provided is the larger of the two deviations of the Clopper-Pearson (\citealp{ClopperPe34}) exact binomial 95\% confidence bounds from the point estimate.
	}
\end{table*}

\begin{table*}
	\centering
	{\small
		\begin{tabular}{@{}c|c@{}c@{}cccc@{}}\toprule
			\textbf{Evaluation}	 &\multicolumn{6}{c}{\textbf{Short Premise}}\\
			\midrule
			\textbf{Fine-tuning (on same)} && \textbf{No} && \multicolumn{3}{c}{\textbf{Yes}}\\
			\cmidrule{1-1} \cmidrule{3-3} \cmidrule{5-7}
			\textbf{Subset} && \textbf{All} && \textbf{All} & \multirow{3}{2cm}{\centering \textbf{Above-Median Long Premise Length}} & \multirow{3}{2cm}{\centering \textbf{Hypothesis Contains Negation}}\\
			&&&&&&\\
			&&&&&&\\
			\midrule
			\multicolumn{7}{c}{[Entail/Neutral/Contradict. Chance=1/3]}\\
			\midrule
			\centering \centering albert-xxlarge-v2\_anli&&0.551+/-0.016&&0.882+/-0.01&0.866+/-0.016&0.885+/-0.015\\
			\centering albert-xxlarge-v2\_ConTRoL-dataset&&0.478+/-0.016&&0.878+/-0.011&0.858+/-0.016&0.881+/-0.015\\
			\centering albert-xxlarge-v2\_vanilla&&0.336+/-0.015&&0.883+/-0.01&0.861+/-0.016&0.883+/-0.015\\
			&&&&&&\\
			\centering allenai\_longformer-base-4096\_anli&&0.402+/-0.015&&0.802+/-0.013&0.767+/-0.019&0.809+/-0.018\\
			\centering allenai\_longformer-base-4096\_ConTRoL-dataset&&0.376+/-0.015&&0.791+/-0.013&0.762+/-0.02&0.793+/-0.019\\
			\centering allenai\_longformer-base-4096\_vanilla&&0.339+/-0.015&&0.779+/-0.013&0.741+/-0.02&0.784+/-0.019\\
			&&&&&&\\
			\centering facebook\_bart-large\_anli&&0.532+/-0.016&&0.879+/-0.011&0.862+/-0.016&0.878+/-0.015\\
			\centering facebook\_bart-large\_ConTRoL-dataset&&0.468+/-0.016&&0.876+/-0.011&0.861+/-0.016&0.879+/-0.015\\
			\centering facebook\_bart-large\_vanilla&&0.323+/-0.015&&0.89+/-0.01&0.874+/-0.016&0.893+/-0.014\\
			&&&&&&\\
			\centering google\_bigbird-roberta-base\_anli&&0.403+/-0.015&&0.84+/-0.012&0.825+/-0.018&0.838+/-0.017\\
			\centering google\_bigbird-roberta-base\_ConTRoL-dataset&&0.383+/-0.015&&0.845+/-0.012&0.824+/-0.018&0.845+/-0.017\\
			\centering google\_bigbird-roberta-base\_vanilla&&0.354+/-0.015&&0.842+/-0.012&0.824+/-0.018&0.841+/-0.017\\
			&&&&&&\\
			\centering nlpaueb\_legal-bert-base-uncased\_anli&&0.514+/-0.016&&0.849+/-0.012&0.829+/-0.018&0.845+/-0.017\\
			\centering nlpaueb\_legal-bert-base-uncased\_ConTRoL-dataset&&0.471+/-0.016&&0.839+/-0.012&0.818+/-0.018&0.833+/-0.017\\
			\centering nlpaueb\_legal-bert-base-uncased\_vanilla&&0.346+/-0.015&&0.853+/-0.011&0.831+/-0.017&0.854+/-0.016\\
			&&&&&&\\
			\centering roberta-large\_anli&&0.374+/-0.015&&0.884+/-0.01&0.867+/-0.016&0.886+/-0.015\\
			\centering roberta-large\_ConTRoL-dataset&&0.478+/-0.016&&0.872+/-0.011&0.85+/-0.017&0.877+/-0.015\\
			\centering roberta-large\_vanilla&&0.4+/-0.015&&0.887+/-0.01&0.868+/-0.016&0.89+/-0.015\\
			&&&&&&\\
			\centering zlucia\_custom-legalbert\_anli&&0.536+/-0.016&&0.843+/-0.012&0.826+/-0.018&0.84+/-0.017\\
			\centering zlucia\_custom-legalbert\_ConTRoL-dataset&&0.462+/-0.016&&0.839+/-0.012&0.821+/-0.018&0.837+/-0.017\\
			\centering zlucia\_custom-legalbert\_vanilla&&0.267+/-0.014&&0.844+/-0.012&0.834+/-0.017&0.848+/-0.017\\
			\midrule
			\centering N&&3966&&3966&1936&1983\\
			\midrule
			\multicolumn{7}{c}{[Entail/Not entail. Chance=1/2]}\\
			\midrule
			\centering 			\centering albert-xxlarge-v2\_DocNLI&&0.508+/-0.014&&0.907+/-0.008&0.896+/-0.012&0.913+/-0.011\\
			\centering allenai\_longformer-base-4096\_DocNLI&&0.496+/-0.014&&0.834+/-0.01&0.813+/-0.015&0.855+/-0.014\\
			\centering facebook\_bart-large\_DocNLI&&0.531+/-0.014&&0.874+/-0.009&0.865+/-0.014&0.889+/-0.013\\
			\centering google\_bigbird-roberta-base\_DocNLI&&0.508+/-0.014&&0.863+/-0.01&0.849+/-0.014&0.875+/-0.013\\
			\centering nlpaueb\_legal-bert-base-uncased\_DocNLI&&0.466+/-0.014&&0.873+/-0.009&0.859+/-0.014&0.89+/-0.013\\
			\centering roberta-large\_DocNLI&&0.412+/-0.013&&0.91+/-0.008&0.897+/-0.012&0.927+/-0.011\\
			\centering zlucia\_custom-legalbert\_DocNLI&&0.616+/-0.013&&0.857+/-0.01&0.843+/-0.014&0.873+/-0.013\\
			\midrule
			\centering N&&5288&&5288&2637&2644\\
			\bottomrule
		\end{tabular}
	}
	\caption{\label{tab:table2b}
		Accuracy on LawngNLI's ``analysis'' subset: Full intermediate-fine-tuned model panel. If fine-tuned, fine-tuning on the LawngNLI subset is on premises with the same granularity as evaluation. The error provided is the larger of the two deviations of the Clopper-Pearson (\citealp{ClopperPe34}) exact binomial 95\% confidence bounds from the point estimate.
	}
\end{table*}

\begin{table*}
	\centering
	{\scriptsize
		\begin{tabular}{@{}p{4.5cm}@{}|@{}p{0.1cm}@{}c@{}c@{}c|@{}p{0.1cm}@{}c@{}c@{}c@{}}\toprule
			\centering \textbf{Needs long premises for fine-tuning}	 &
			&\multicolumn{3}{c|}{\centering \textbf{No}}&
			&\multicolumn{3}{c}{\centering \textbf{Yes}}\\
			\midrule
			\multirow{2}{4.5cm}{\centering \textbf{Fine-tuning}}	 &
			&\multicolumn{3}{c|}{\multirow{2}{1.8cm}{\centering \textbf{BM25 retrieval on short premise}}}&
			&\multicolumn{3}{c}{\multirow{2}{1.8cm}{\centering \textbf{Long premise filtered by BM25 retrieval}}}\\
			&  &  &  &  &  &  &  &\\
			\midrule
			\multirow{2}{4.5cm}{\centering \textbf{Long premise filtered by BM25 retrieval at evaluation}}	 &
			&\multicolumn{3}{c|}{\multirow{2}{1.8cm}{\centering \textbf{Yes}}}&
			&\multicolumn{3}{c}{\multirow{2}{1.8cm}{\centering \textbf{Yes}}}\\
			&  &  &  &  &  &  &  &\\
			\midrule
			\centering \textbf{Label}&\phantom{abc}  & \textbf{Automatic} &\phantom{abc}  & \textbf{Gold} &\phantom{abc}  & \textbf{Automatic} &\phantom{abc}  & \textbf{Gold}\\
			\cmidrule{1-1} \cmidrule{3-3} \cmidrule{5-5} \cmidrule{7-7} \cmidrule{9-9}
			&  & \textbf{(4)} &  & \textbf{(4)} &  & \textbf{(6)} &  & \textbf{(6)}\\
			\midrule
			\multicolumn{9}{c}{[Entail/Neutral/Contradict. Chance=1/3]}\\
			\midrule
\centering albert-xxlarge-v2\_anli&&0.668&&0.688&&0.684&&0.689\\
\centering albert-xxlarge-v2\_ConTRoL-dataset&&0.674&&0.673&&0.629&&0.65\\
\centering albert-xxlarge-v2\_vanilla&&0.685&&0.677&&0.646&&0.636\\
\centering &&&&&&&&\\
\centering facebook\_bart-large\_anli&&0.558&&0.537&&0.683&&0.68\\
\centering facebook\_bart-large\_ConTRoL-dataset&&0.614&&0.604&&0.672&&0.67\\
\centering facebook\_bart-large\_vanilla&&0.622&&0.583&&0.672&&0.699\\
			\midrule
			\centering N &  \multicolumn{8}{c}{140} \\
			\bottomrule
		\end{tabular}
	}
	\caption{\label{tab:table4}
		Balanced accuracy/macro-averaged recall: Top 6 three-label models from RQ1 (Appendix Table~\ref{tab:table32a}), evaluated against the automatic labels versus gold labels on the human-validated subset. Column numbers are the same as in Appendix Table~\ref{tab:table32a}. As described in Section~\ref{sec:goldlabels}, for the gaps between the top model setups that need long premises for fine-tuning (column 6) versus those that do not (column 4), there are non-positive biases (automatic labels versus gold labels) here both for the model with the largest performance gap from Table~\ref{tab:table32} and on average across the 6 models. Thus this comparison provides evidence that the performance gap from RQ1 is not due to bias from error in the automatic labels.
	}
\end{table*}

\begin{table*}
	\centering
	{\scriptsize
		\begin{tabular}{@{}p{4.5cm}@{}|@{}p{0.1cm}ccccccc}\toprule
			\multirow{3}{4.5cm}{\centering \textbf{Evaluation}}&\phantom{abc}  &\multicolumn{3}{@{}c}{\multirow{3}{2.2cm}{\centering \textbf{Long premise}}}  &  &\multicolumn{3}{@{}c}{\multirow{3}{2.2cm}{\centering \textbf{Long premise filtered by BM25 retrieval}}}\\
			&  &  &  &  &  &  &  &\\
			&  &  &  &  &  &  &  &\\
			\cmidrule{1-1} \cmidrule{3-5} \cmidrule{7-9}
			\centering \textbf{Recall@k}&&\textbf{1}&\textbf{10}&\textbf{100}&&\textbf{1}&\textbf{10}&\textbf{100}\\
			\midrule
			\multicolumn{9}{c}{[Entail hypotheses as queries]}\\
			\midrule
\centering all-distilroberta-v1&&0.039+/-0.012&0.108+/-0.018&0.262+/-0.025&&0.075+/-0.016&0.219+/-0.023&0.505+/-0.027\\
\centering all-mpnet-base-v2&&0.042+/-0.012&0.115+/-0.018&0.224+/-0.023&&0.083+/-0.016&0.219+/-0.023&0.433+/-0.027\\
\centering nli-distilroberta-base-v2&&0.005+/-0.006&0.035+/-0.011&0.132+/-0.019&&0.014+/-0.008&0.063+/-0.014&0.243+/-0.024\\
\centering msmarco-distilroberta-base-v2&&0.02+/-0.009&0.053+/-0.013&0.152+/-0.021&&0.045+/-0.013&0.14+/-0.02&0.359+/-0.027\\
\centering BM25&&0.138+/-0.02&0.281+/-0.025&0.447+/-0.027&&0.158+/-0.021&0.308+/-0.026&0.458+/-0.027\\
			\midrule
			\multicolumn{9}{c}{[Contradict hypotheses as queries]}\\
			\midrule
\centering all-distilroberta-v1&&0.037+/-0.012&0.1+/-0.017&0.263+/-0.025&&0.069+/-0.015&0.22+/-0.023&0.486+/-0.027\\
\centering all-mpnet-base-v2&&0.039+/-0.012&0.112+/-0.018&0.222+/-0.023&&0.078+/-0.016&0.215+/-0.023&0.43+/-0.027\\
\centering nli-distilroberta-base-v2&&0.006+/-0.006&0.03+/-0.011&0.13+/-0.019&&0.01+/-0.007&0.063+/-0.014&0.238+/-0.024\\
\centering msmarco-distilroberta-base-v2&&0.018+/-0.009&0.054+/-0.014&0.151+/-0.02&&0.04+/-0.012&0.143+/-0.02&0.359+/-0.027\\
\centering BM25&&0.141+/-0.02&0.267+/-0.025&0.429+/-0.027&&0.145+/-0.02&0.286+/-0.025&0.452+/-0.027\\
			\midrule
			\centering N &  \multicolumn{8}{c}{1322} \\
			\bottomrule
		\end{tabular}
	}
	\caption{\label{tab:table34}
		Recall@k of SentenceTransformers (\citealp{ReimersGu21}) bi-encoders and BM25 (\citealp{RobertsonZa09}) baseline evaluated on implication-based retrieval (of each LawngNLI test set premise case using hypothesis as query, among 999 other cases similar to the query; see Section~\ref{sec:evaluation2}), using a dataset derived from LawngNLI. The error provided is the larger of the two deviations of the Clopper-Pearson (\citealp{ClopperPe34}) exact binomial 95\% confidence bounds from the point estimate.
	}
\end{table*}

\end{document}